\documentclass[article,12pt]{elsarticle}
\usepackage{setspace}
\usepackage{amssymb}
\usepackage{url}
\usepackage{graphics}
\usepackage{graphicx}
\usepackage{amsmath}
\usepackage{amsfonts}
\usepackage{mathtools}
\usepackage{nicefrac}
\usepackage{caption}
\usepackage{subcaption}
\usepackage{multirow}
\begin{document}
	\begin{frontmatter}
		
		\title{Feature selection simultaneously preserving both class and cluster structures}
		
		\author{Suchismita Das\corref{mycorrespondingauthor} and  Nikhil R. Pal}
		\cortext[mycorrespondingauthor]{Corresponding author}
		\address{Electronics and Communication Sciences Unit, Indian Statistical Institute, 203 B T Road, Kolkata-700108}
		\ead{suchismitasimply@gmail.com,nikhil@isical.ac.in}
		
		%
		%
		
		\begin{abstract}
		When a data set has significant differences in its class and cluster structure, selecting features aiming only at the discrimination of classes would lead to poor clustering performance, and similarly, feature selection aiming only at preserving cluster structures would lead to poor classification performance. To the best of our knowledge, a feature selection method that simultaneously considers class discrimination and cluster structure preservation is not available in the literature. In this paper, we have tried to bridge this gap by proposing a neural network-based feature selection method that focuses both on class discrimination and structure preservation in an integrated manner. In addition to assessing typical classification problems, we have investigated its effectiveness on band selection in hyperspectral images.  Based on the results of the experiments, we may claim that the proposed feature/band selection can select a subset of features that is good for both classification and clustering.
		\end{abstract}
		
		
	\end{frontmatter}
	
	
	\section{Introduction}\label{sec:chap4_intro}

	Feature selection methods can be broadly classified on the basis of the utilization of the class label information. There are three categories: supervised, semi-supervised and unsupervised \cite{zhang2019feature,solorio2020review}. The supervised feature selection method exploits the label information to find out the relevant features which distinguish samples of different classes \cite{atashgahi2023supervised,jiao2022solving}. Semi-supervised feature selection is used when some labeled samples along with  plenty of unlabelled samples are present \cite{an2023robust,zeng2019local}. Both labeled and unlabelled data are used to modify a hypothesis obtained from the labeled data \cite{chandrashekar2014survey}. Unsupervised feature selection is much more difficult as it needs to find out the useful features in the absence of the label information \cite{solorio2020review,wang2015unsupervised}. Different criteria have been chosen to select a subset of original features in different unsupervised feature selection studies. Some of them are: preserving the data distribution such as manifold structure \cite{du2015unsupervised,wu2021joint}, preserving cluster structure \cite{li2013clustering,hou2013joint}, and preserving data similarity \cite{zhao2011similarity,gao2021preserving}. It is noteworthy that in the case of unsupervised feature selection, some methods try to preserve the ``structure" or ``geometry" of the data in some sense. Contrarily supervised feature selection methods in most cases do not set any explicit criteria to preserve the structure of the data. They only pay heed to separating the classes as much as possible with different measures exploiting class information such as  Fisher score \cite{gu2012generalized, song2017feature}, Laplacian score \cite{he2005laplacian}, mutual information \cite{peng2005feature,vergara2014review}, normalized mutual information \cite{peng2005feature,estevez2009normalized,tesmer2004amifs,vergara2010cmim}, ReliefF \cite{urbanowicz2018relief}, class correlation \cite{wang2016supervised}, classifier score \cite{le2020novel}. We should note here that the feature selection criterion are not always lead by a single objective. Feature selection methods often follow a criterion that consisits of two or more objectives. The study in \cite{wang2015unsupervised} proposes a criterion named `maximum projection and minimum redundancy' which is governed by two goals: projecting data into a feature subspace with minimum reconstruction error and minimum redundancy. The studies in \cite{zhou2016global,zhu2018adaptive,liu2013global,zhu2017local} claim that both global structure and local structure should be preserved in the projected space as both them may carry important discriminating information and hence, they  have proposed feature selection schemes that focus both on global and local structure preservation. The investigation in \cite{ye2020dual} claims to preserve dual global structures.  Going through various feature selection schemes having multiple objective we found that whenever class label is available, no work in feature selection explicitly focused preserving structural information along with class information although both of these are important discriminative information and may have positive impact on the generalization ability of the classifier. Suppose, for a data set, the class and cluster structures are substantially different. Exploiting only the class labels, it may not be possible to keep the cluster structures in the projected space. For a practical system, even when the primary task is classification, we may need to cluster the samples in the space defined by the selected features. For example, fuzzy rule based classifiers are often designed by clustering the training data for each class and translating each cluster into a rule\cite{chakraborty2004neuro,chen2012integrated,xue2022adaptive}. We could not find any feature selection method that focuses both on class and cluster separability. To bridge this gap, in this study we propose a feature selection method that selects features preserving class and cluster-structure information simultaneously. We employ a multi-layer perceptron (MLP) based neural network to develop an embedded feature selection scheme. The training of the proposed MLP based feature selection method is governed by both class discriminating and cluster (structure) preserving objectives. The philosophy is quite general and can be easily extended to other networks such as radial basis function network.
	
	\section{Proposed Method}\label{sec:chap4_proposed}
\begin{figure}[!hb]
		\centering
		\includegraphics[width=0.5\linewidth]{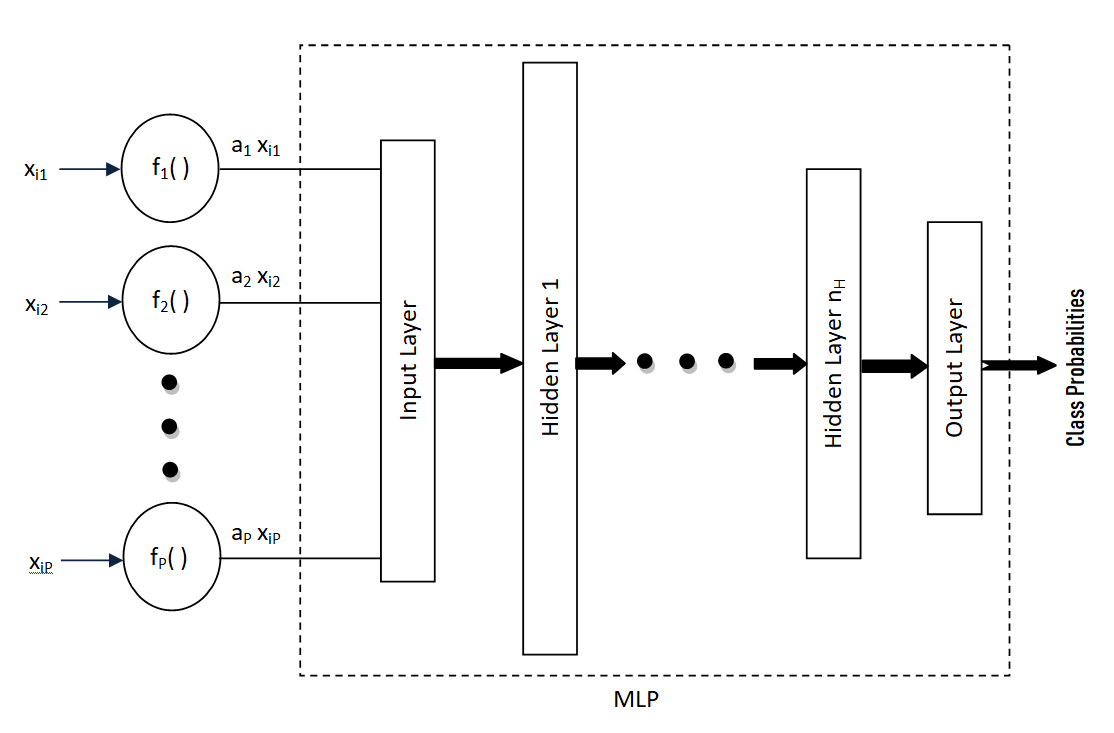}
		\caption{Proposed network for feature selection}
		\label{fig:network1}
	\end{figure}

	Let us denote the input data by an $n\times P$ matrix, $\mathbf{X}=\{\mathbf{x}_{i} \in \mathbb{R}^{P}\}_{i=1}^{n}$. Here, $\mathbf{x}_{i}$ is a $P$ dimensional row vector of the form, $\mathbf{x}_{i}=(x_{i1},x_{i2},\cdots,x_{iP})$.   The collection of class labels of  $\mathbf{X}$ be $\mathbf{Z}=\{z_{i} \in\{1,2,\cdots, C\}\}_{i=1}^{n}$, where, $z_{i}$ is the class label corresponding to $\mathbf{x}_{i}$. We aim to select a subset of size $Q$ from the original set of features such that the selected subset performs reasonably well in terms of the classification task as well as in clustering. In other words, if we design a classifier using the selected features, the performance of the classifier would be comparable to a classifier designed using all features. Similarly, if we cluster the data in the reduced dimension as well as in the original dimension we expect to get similar partition matrix. Here, we propose a neural network-based framework to select features. Neural networks have been explored for the feature selection \cite{setiono1997neural,castellano2000variable,verikas2002feature} as well as for classification \cite{zhang2000neural,dreiseitl2002logistic,agrawal2015neural,mabrouk2020deep}. However, in our proposed model the neural network simultaneously selects features and learns a classifier as we follow an embedded method for feature selection. Moreover, our proposed network preserves structural information and the class label  information simultaneously, whereas, the feature selection networks in \cite{setiono1997neural,verikas2002feature} solve  classification problems, consider class label information in their loss function but not any structural information. Note here, the work in \cite{castellano2000variable} considers a system identification problem. To build the neural network-based embedded feature selector, we employ the multi-layer perceptron (MLP) based framework used in \cite{chakraborty2008selecting,chakraborty2014feature,chakraborty2014sensor}. The basic framework is shown in Fig. \ref{fig:network1}.
	As seen in Figure \ref{fig:network1}, preceding the input layer of the MLP, there is a layer consisting of $P$ nodes. Before entering to the input layer of the MLP, the $j$\textsuperscript{th} feature passes through the node $f_{j}()$.  These nodes act as attenuating gates that allow or block features from contributing to the output of the neural network effectively. For the $i$\textsuperscript{th} instance, it's $j$\textsuperscript{th} feature $x_{ij}$ on passing the gate node $f_{j}()$ becomes $a_{j}x_{ij}$; i.e., $f_{j}(x_{ij})=a_{j}x_{ij}$. In MLP, a weighted sum of the values available at the input nodes is applied to the hidden nodes of the first hidden layer. Zero value at an input node implies that the corresponding feature is not considered. When training of the MLP-based framework is complete, $a_{j}$s for the selected features become close to $1$, effectively allowing them to contribute to the classifier. Whereas, for poor or rejected features $a_{j}$s become close to $0$, effectively making them not contribute to the classifier. In \cite{chakraborty2008selecting,chakraborty2014feature,chakraborty2014sensor}, this framework was explored for classification-oriented feature selection, group feature selection, and redundancy-controlled feature selection. Here, we explore this framework for  simultaneous structure-preserving and class-discriminating feature selection. Next, we elaborate on the MLP-based framework and the proposed objective functions to train the network. 
	
	We denote the $P$ nodes before the input layer of the MLP as  $f_{j}()$s for $j=1,2,\dots P$ where $f_{j}()$ is a gate or modulator function applied on the $j$\textsuperscript{th} feature, $x_{j}$. Now, we have to design $f_{j}()$ in such a way,
	\begin{equation}\label{eq:gate_func}
		f_{j}(x_{j})=a_{j}x_{j}=\begin{cases}
			{x}_{j} &\text{ if }{x}_{j}\text{ is a useful feature.}\\
			0 &\text{ otherwise. }
		\end{cases}
	\end{equation}
	In our framework, the factor, $a_{j}$ is learnable. We implement $a_{j}$ as a smooth continuous function, $a_{j}=\exp(-\lambda_{j}^{2})$. Clearly, when $\lambda_{j}=0$, the value of $\exp(-\lambda_{j}^{2})= 1$ and when $\lambda_{j} \rightarrow \pm \infty$, the value of $\exp(-\lambda_{j}^{2})=0$. By adding suitable regularizer terms to the objective function we design our learning system in such a way that, over the learning process, the gate parameters, $\lambda_{j}$s for  useful features drop close to zero and that for derogatory or indifferent features rise to high values. So, in our learning system, the learnable parameters, $\lambda_{j}$s and the  neural network weights are learned together, i.e., the loss function is minimized with respect to both $\lambda_{j}$s and the  neural network weights. 
	
	Now, we have to define a suitable loss function for selecting features along with learning the embedded classifier. Our aim is to select features that are reasonably good for classification as well as clustering. To satisfy this requirement, we take the loss function as a combination of two losses $E_{class}$ and $E_{struct}$.  $E_{class}$ is considered for preserving class information and $E_{struct}$ is considered for preserving structural information. At this moment let us consider the network for selecting features for efficient classification only. A suitable loss function to impose class discrimination is the cross-entropy loss \cite{zhang2018generalized}. We define, $E_{class}$ as the cross-entropy loss involving actual and predicted class labels. 	
	\begin{equation}\label{eq:class_error}
		E_{class}=-\dfrac{1}{n}\sum_{i=1}^{n}\sum_{k=1}^{C}t^{i}_{k}\log\left(p_{k}(\mathbf{x}_{i}) \right) 
	\end{equation}
	Here, $t^{i}_{k}$ is $k$\textsuperscript{th} element of the one-hot encoded label of the sample $\mathbf{x}_{i}$ or in other words  $t^{i}_{k}$ is the $k$\textsuperscript{th} element of the vector $\mathbf{t}^{i}\in \left\lbrace 0,1\right\rbrace^{C} $ such that
	\begin{equation}
		t_{k}^{i}= \begin{cases}
			1 &\text{ if } k=z_{i}\\
			0 &\text{ otherwise}
		\end{cases}
	\end{equation}
	In (\ref{eq:class_error}), $p_{k}(\mathbf{x}_{i})$ is the predicted probability (by the MLP classifier) of  $\mathbf{x}_{i}$ being in $k$\textsuperscript{th} class. As already discussed above, for effective feature selection, the magnitude of $\lambda_{j}$s for the selected features should drop to almost zero and for rejected features should rise to high values. To ensure this condition we add the following regularizer.
	\begin{align}\label{eq:effect_select}
		E_{select} = &\dfrac{1}{P}\sum_{j=1}^{P} a_{j}(1-a_{j}) \nonumber\\
		= &\dfrac{1}{P}\sum_{j=1}^{P} \exp(-\lambda_{j}^{2})(1-\exp(-\lambda_{j}^{2})) 
	\end{align}
	In a feature selection framework, a constraint for selecting a fixed number of features is necessary. The following regularizer tries to keep the number of the selected features close or equal to $Q$.
	\begin{equation}\label{eq:num_bands}
		E_{Q}=\dfrac{1}{Q^{2}}\{(\sum_{j=1}^{P}a_{j})-Q\}^{2}=\dfrac{1}{Q^{2}}\{(\sum_{j=1}^{P}\exp(-\lambda_{j}^{2}))-Q\}^{2}
	\end{equation}
	So, the overall loss function for the selection of features with our framework for classification purposes is the following.
	\begin{equation}\label{eq:error_cl_band}
		E= E_{class} + \alpha_{1}E_{select} + \alpha_{2}E_{Q}
	\end{equation}
	Here, $\alpha_{1} \ge 0,\alpha_{2} \ge 0$ are scalar multipliers for adjusting the importance of $E_{select}$ and $E_{Q}$ in the overall error function $E$.
	
	Now let us focus on our original agenda of selecting features that perform satisfactorily both for classification and clustering. To preserve structural information of the data in the lower dimensional space formed by the selected $Q$ features, we consider the Sammon's stress \cite{sammon1969nonlinear} as a loss function. The Sammon's stress is the loss function for a non-linear mapping named Sammon's mapping that is able to capture complex non-linear structures in data, as a result, also preserves cluster structure.  The lower the value of Sammon's stress, the better the lower dimensional representations in capturing the original inter-point distances or structures of the original data. We can define Sammon's stress involving the original input space and selected feature space as the following. 
	\begin{equation}\label{eq:sammons}
		E_{sammons}=\dfrac{1}{(\sum_{i,l=1}^{n} d_{il})} \sum_{i=1}^{n-1} \sum_{l=i+1}^{n} \dfrac{\left( d_{il}^{\mathbf{X}}- d_{il}^{\mathbf{\hat{X}}}\right)^{2}}{d_{il}^{\mathbf{X}}}
	\end{equation}
	$d_{il}^{\mathbf{X}}$ is the distance between $\mathbf{x}_{i}$ and $\mathbf{x}_{l}$. $\mathbf{\hat{X}}=\{\mathbf{\hat{x}}_{i}=(a_{1}x_{i1},a_{2}x_{i2},\cdots,a_{P}x_{iP})^{T} \in \mathbb{R}^{P}\}_{i=1}^{n}$. So,	$d_{il}^{\mathbf{\hat{X}}}$ is the distance between $\mathbf{\hat{x}}_{i}$ and $\mathbf{\hat{x}}_{l}$. As discussed earlier, at the end of the training of our embedded system, $a_{j}$s will be close to $0$ or $1$ depending on whether the corresponding features are rejected or selected. Therefore, for a trained system $d_{il}^{\mathbf{\hat{X}}}$ would signify the distance between $i$\textsuperscript{th} and $l$\textsuperscript{th} instances in the latent space formed by the implicitly selected $Q$ features. So considering $E_{sammons}$ in Equation (\ref{eq:sammons}) as an regularizer, the resultant overall loss function is given by.
	\begin{equation}\label{eq:overall_error_nonlarge_n}
		E_{tot}=E_{class}+ \beta E_{sammons} + \alpha_{1} E_{select} + \alpha_{2} E_{Q}
	\end{equation}
	$\beta \ge 0$ is a scalar multiplier that controls the trade-off between the class information and the structural information in the feature selection process.
	Note that, the computational complexity for the loss function in Equation (\ref{eq:sammons}) is $O(n^{2})$. For large $n$, computing  Equation (\ref{eq:sammons}) and hence Equation (\ref{eq:overall_error_nonlarge_n}) is intensive. As the weight update at each iteration will involve computing Equation (\ref{eq:sammons}), the overall computation cost would be high. For small and moderate $n$, we use Equation (\ref{eq:overall_error_nonlarge_n}) as the loss function to be minimized. However, for large $n$ to avoid the high computational cost we modify Equation (\ref{eq:sammons}) as follows.
	\begin{equation}\label{eq:struct}
		E_{struct}= \dfrac{1}{(\sum_{\mathbf{x}_{i},\mathbf{x}_{l} \in S_{t}}  d_{il})}\sum_{\mathbf{x}_{i} \in S_{t}} \sum_{\mathbf{x}_{l} \in S_{t}; \mathbf{x}_{l} \ne \mathbf{x}_{i}} \dfrac{\left( d_{il}^{\mathbf{X}}- d_{il}^{\mathbf{\hat{X}}}\right)^{2}}{d_{il}^{\mathbf{X}}}
	\end{equation}    
	Here $S_{t}$ is a randomly selected subset of $\mathbf{X}$ at the $t$\textsuperscript{th} iteration. Different $S_{t}$s are chosen at different iterations and hence different sets of inter-point distances are preserved. Since the considered MLP is trained over a large number of iterations, the use of Equation  (\ref{eq:struct}) is expected to result in almost the same effect as that by Equation (\ref{eq:sammons}). We have to choose $|S_{t}|$ such that Equation (\ref{eq:struct}) is computationally manageable and at the same time it should be large enough to make $E_{struct}$ an effective substitute of $E_{sammons}$. Adding Equation (\ref{eq:struct}) to Equation (\ref{eq:error_cl_band})  we propose the following loss function for our system.
	\begin{equation}\label{eq:overall_error}
		E_{tot}=E_{class}+ \beta E_{struct} + \alpha_{1} E_{select} + \alpha_{2} E_{Q}
	\end{equation}
	$E_{tot}$ is minimized with respect to the gate parameters $\lambda_{j}$s and the weights of the network to find their optimal values.
	
	\section{Experimentation and Results}\label{sec:chap4_expt}
	The feature selection framework proposed in this chapter is generic but it can be adapted to solve specialized problems. We have studied the proposed framework for general datasets as well as for solving a special problem: band selection of hyperspectral images. We present the results of band selection for HSIs in a different subsection, Subsec. \ref{subsec:chap4_expt_HSI}. We present the results of feature selection for the conventional classification problem in the following subsection (Subsec. \ref{subsec:chap4_expt_general}).
	\subsection{Feature selection for conventional classification problems }\label{subsec:chap4_expt_general}
	We have used five publicly available datasets that are very commonly used for classification and clustering. The first four datasets are downloaded from UCI machine learning repository \cite{dua2019}. AR10P is downloaded from the open-source feature selection repository of Arizona State University\cite{li2018feature}. We have also performed the experiments with three benchmark HSI datasets for land cover classification problems. We discuss them in a separate subsection (Subsec. \ref{subsec:chap4_expt_HSI}). 
	
	The details of the number of features, number of classes, and number of instances for the five datasets are summarized in Table \ref{tab:sum_data}. 
	\begin{table}[!ht]
		\caption{Summary of datasets.}\label{tab:sum_data}
		\centering
		\begin{tabular}{|c| c| c| c|}
			\hline
			Name & Number of features & Number of classes & Number of instances \\
			& $P$ & $C$ & $n$ \\
			\hline
			E. coli & 7 & 8 & 336 \\
			Glass & 9 & 6 & 214 \\
			Ionosphere & 34 & 2 & 351\\
			Sonar & 60 & 2 & 208\\
			AR10P & 2400 & 10 & 130\\
			\hline
		\end{tabular}
	\end{table}
	The datasets are used directly without any further processing. The datasets are partitioned into training and test sets as approximately $90\%$ and $10\%$ of the total number of instances. To implement our proposed feature selection scheme we use the neural network shown in Fig. \ref{fig:network1} with the number of hidden layers, $n_H = 1$. The input and output layers have $P$ and $C$ nodes respectively, where $P$ is the number of features and $C$ is the number of classes corresponding to the considered dataset. The number of hidden nodes in the hidden layer is $8$ ($20$ for AR10P data set). To get stable feature selection results, the network weights are initialized in a certain way. To set the initial weights of the proposed network, we undergo the following steps. First, we consider the usual MLP part of our network (i.e. without feature selection), depicted by the portion within the dotted rectangle in Fig. \ref{fig:network1}, and initialize its weights randomly. Next, we train the usual MLP with the cross-entropy loss defined in Equation (\ref{eq:class_error}) with the training set until convergence. The weights of the converged network are used as the initial weights of the proposed network.   The gate parameter $\lambda_{j}$s are initialized with values drawn  randomly from a normal distribution with mean $=2$ and spread $=\frac{1}{\sqrt{P}}$. The initial values of $\lambda_{j}$s are chosen around $2$ to effectively make the gates almost closed initially. As the learning progresses the $\lambda_{j}$s are updated in a way to allow the useful features to the network. For the proposed system, to select a subset of $Q$ features, the gate parameters $\lambda_{j}$s are sorted in ascending order, and the $Q$ features corresponding to the top $Q$, $\lambda_{j}$s are selected. The network weights as well as the gate parameters $\lambda_{j}$s are learned using the adaptive gradient algorithm, \textit{`train.AdagradOptimizer'} routine of the \textit{`TensorFlow'} framework \cite{abadi2016tensorflow}. For all experiments with the data sets in Table \ref{tab:sum_data}, both $\alpha_{1}$ and $\alpha_{2}$ of the error functions in Equations (\ref{eq:error_cl_band}) and (\ref{eq:overall_error_nonlarge_n}) are set as $1$. The total number of iterations for training the network is set to $20000$. The five datasets we consider here, have the number of instances $n<400$, which is not so large. Therefore, we  use (\ref{eq:overall_error_nonlarge_n}) as the overall loss function to train the MLP based architecture for selecting features that are reasonably good for clustering and classification. When $\beta=0$ in (\ref{eq:overall_error_nonlarge_n}), effectively, the error function that governs the learning of our MLP based embedded feature selection scheme is (\ref{eq:error_cl_band}).  The corresponding feature selection scheme now only considers classification. Let us name this method as feature selection with MLP (FSMLP). When $\beta \neq 0$ in (\ref{eq:overall_error_nonlarge_n}), our method takes structure preservation into account along with classification. Let us name the corresponding method as FSMLP\textsubscript{struct}. To understand the importance of adding the structure preserving regularizer (\ref{eq:sammons}), we perform feature selection with FSMLP and compare with FSMLP\textsubscript{struct} having different $\beta$ values. We explore three values of $\beta$s $0.1, 1,$ and $10$. Although the exact value of the $\beta$ that is optimum for a particular dataset for a particular number of selected features $Q$ cannot be decided from these three values, we investigate the effect of three widely different $\beta$s to see the role of the weight to the structure preserving regularizer, i.e. $\beta$ on the performance of the selected features. We compare with three other methods namely, Independent Component Analysis (ICA)-based feature selection \cite{prasad2004efficient},  F-score based filter method \cite{song2017feature}, and  mutual information based filter method \cite{peng2005feature,vergara2014review}. The performance of both FSMLP and FSMLP\textsubscript{struct} is dependent on the initial weights of the network. So, we repeat the initialization of the network weights and gate parameters $\lambda_{j}$s five times and run the schemes- FSMLP or FSMLP\textsubscript{struct} five times with the five initializations. For the performance measure of FSMLP and  FSMLP\textsubscript{struct}, we consider the  average performance over the five subsets obtained from the five runs.  To check the effectiveness of the methods in selecting features that perform well in classification and clustering simultaneously, we compute the classification scores of the support vector machine (SVM) classifier as well as several structure-preserving indices: Sammon's stress (SS) \cite{sammon1969nonlinear}, normalized mutual information (NMI) \cite{peng2005feature,estevez2009normalized,tesmer2004amifs,vergara2010cmim}, adjusted rand index (ARI) \cite{steinley2004properties}, and Jaccard Index (JI) \cite{fletcher2018comparing}. As the measure of classification performance, we use the overall classification accuracy (OCA) of  the SVM classifier. The optimal hyper-parameters of SVM are determined through five-fold cross-validation using grid search. Note that here the test set is not only unseen to the SVM classifier but unseen to the feature selection methods also. SS, defined in Equation (\ref{eq:sammons}) use the original inter-point distances $d_{il}^{\mathbf{X}}$s and latent space inter-point distances $d_{il}^{\mathbf{\hat{X}}}$s. Here to compute $d_{il}^{\mathbf{\hat{X}}}$, we use the lower dimensional data formed by the selected $Q$ features. We use NMI, ARI, and JI as the structure-preserving performance metrics by supplying the cluster labels obtained from clustering the data in the original space (using all features) as the true label and the cluster labels obtained from clustering the data in the reduced space formed by the selected $Q$ features as the predicted cluster label. So, NMI, ARI, and JI measure how the cluster assignments in the original space and in the selected space agree, effectively giving a measure for the preservation of the original cluster structure in the selected space. We know that the maximum value for NMI or ARI or JI is $1$. Here, the value of each of these three measures being close to $1$ indicates that the cluster structure in the original space is preserved in the selected space. As the clustering algorithm we use, the fuzzy $C$ means  (FCM) algorithm \cite{bezdek1984fcm} with the fuzzy exponent $m=2$. We set the number of clusters for FCM algorithm  as the number of classes. We use two values for the number of the selected features, $Q$. $Q=0.35 \times P$ and $Q=0.5 \times P$, where these values are rounded up to the nearest integers using the ceiling function.

	Tables \ref{tab:perf_ecoli_train} and \ref{tab:perf_ecoli_test} summarize the performances of the proposed method and other comparing methods for training and test sets, respectively for the E. coli dataset.  We tabulate the three previously mentioned structure preserving measures and one classifier score for two choices of the number of selected features (approximately $35\%$ and $50\%$  of the original dimension) i.e., $Q=3$, and $Q=4$ in Tables \ref{tab:perf_ecoli_train} and \ref{tab:perf_ecoli_test}. 
	\begin{table}[!tb]
		\caption{Performance comparison for E. coli, training set for different choices of $\beta$ and $Q$.}\label{tab:perf_ecoli_train}
		\centering
			\resizebox{12.5cm}{!}{
			\begin{tabular}{|c|c c c c c c c c c c|}
				\hline
				Method	&	\multicolumn{2}{c}{SS}			&	\multicolumn{2}{c}{NMI}			&\multicolumn{2}{c}{ARI}			&	\multicolumn{2}{c}{JI}			&	\multicolumn{2}{c|}{OCA}			\\
				&	Q=3	&	Q=4	&	Q=3	&	Q=4	&	Q=3	&	Q=4	&	Q=3	&	Q=4	&	Q=3	&	Q=4	\\
				\hline
				ICA	&	0.2323	&	0.2146	&	0.4250	&	0.4220	&	0.2643	&	0.2617	&	0.4250	&	0.4220	&	0.6788	&	0.6656	\\
				F Score	&	0.1547	&	0.0648	&	0.5417	&	0.6124	&	0.3520	&	0.5073	&	0.5417	&	0.6124	&	0.6325	&	0.7815	\\
				Mutual Info	&	0.0726	&	0.0204	&	0.6061	&	0.7511	&	0.4993	&	0.6884	&	0.6061	&	0.7511	&	0.6823	&	0.6719	\\
				FSMLP	&	0.1391	&	0.1098	&	0.4356	&	0.4751	&	0.2705	&	0.3167	&	0.4356	&	0.4751	&	0.8391	&	0.8828	\\
				FSMLP\textsubscript{struct},	&	\multirow{ 2}{*}{0.1391}	&	\multirow{ 2}{*}{0.1098}	&	\multirow{ 2}{*}{0.4356}	&	\multirow{ 2}{*}{0.4751}	&	\multirow{ 2}{*}{0.2705}	&	\multirow{ 2}{*}{0.3167}	&	\multirow{ 2}{*}{0.4356}	&	\multirow{ 2}{*}{0.4751}	&	\multirow{ 2}{*}{0.8391}	&	\multirow{ 2}{*}{0.8828}	\\
				$\beta=0.1$	&		&		&		&		&		&		&		&		&		&		\\
				FSMLP\textsubscript{struct},	&	\multirow{ 2}{*}{0.0902}	&	\multirow{ 2}{*}{0.0391}	&	\multirow{ 2}{*}{0.5338}	&	\multirow{ 2}{*}{0.6138}	&	\multirow{ 2}{*}{0.3868}	&	\multirow{ 2}{*}{0.4873}	&	\multirow{ 2}{*}{0.5338}	&	\multirow{ 2}{*}{0.6138}	&	\multirow{ 2}{*}{0.9000}	&	\multirow{ 2}{*}{0.9007}	\\
				$\beta=1$	&		&		&		&		&		&		&		&		&		&		\\
				FSMLP\textsubscript{struct},	&	\multirow{ 2}{*}{0.0766}	&	\multirow{ 2}{*}{0.0280}	&	\multirow{ 2}{*}{0.5812}	&	\multirow{ 2}{*}{0.7137}	&	\multirow{ 2}{*}{0.4497}	&	\multirow{ 2}{*}{0.6362}	&	\multirow{ 2}{*}{0.5812}	&	\multirow{ 2}{*}{0.7137}	&	\multirow{ 2}{*}{0.8675}	&	\multirow{ 2}{*}{0.8768}	\\
				$\beta=10$	&		&		&		&		&		&		&		&		&		&		\\
				\hline
		\end{tabular}}
	\end{table}
	\begin{table}[!tb]
		\caption{Performance comparison for E. coli, test set for different choices of $\beta$ and $Q$.}\label{tab:perf_ecoli_test}
		\centering
		\resizebox{12.5cm}{!}{
			\begin{tabular}{|c|c c c c c c c c c c|}
				\hline
				Method	&	\multicolumn{2}{c}{SS}			&	\multicolumn{2}{c}{NMI}			&\multicolumn{2}{c}{ARI}			&	\multicolumn{2}{c}{JI}			&	\multicolumn{2}{c|}{OCA}			\\
				&	Q=3	&	Q=4	&	Q=3	&	Q=4	&	Q=3	&	Q=4	&	Q=3	&	Q=4	&	Q=3	&	Q=4	\\
				\hline
				ICA	&	0.2562	&	0.2332	&	0.6191	&	0.6068	&	0.2858	&	0.2775	&	0.6191	&	0.6068	&	0.5588	&	0.6471	\\
				F Score	&	0.1389	&	0.0682	&	0.7124	&	0.7150	&	0.4579	&	0.4552	&	0.7124	&	0.7150	&	0.5882	&	0.6471	\\
				Mutual Info	&	0.0763	&	0.0426	&	0.7047	&	0.8047	&	0.4541	&	0.6114	&	0.7047	&	0.8047	&	0.4545	&	0.5909	\\
				FSMLP	&	0.1217	&	0.0976	&	0.7341	&	0.7416	&	0.5284	&	0.5449	&	0.7341	&	0.7416	&	0.6824	&	0.7000	\\
				FSMLP\textsubscript{struct},	&	\multirow{ 2}{*}{0.1217}	&	\multirow{ 2}{*}{0.0976}	&	\multirow{ 2}{*}{0.7341}	&	\multirow{ 2}{*}{0.7416}	&	\multirow{ 2}{*}{0.5284}	&	\multirow{ 2}{*}{0.5449}	&	\multirow{ 2}{*}{0.7341}	&	\multirow{ 2}{*}{0.7416}	&	\multirow{ 2}{*}{0.6824}	&	\multirow{ 2}{*}{0.7000}	\\
				$\beta=0.1$	&		&		&		&		&		&		&		&		&		&		\\
				FSMLP\textsubscript{struct},	&	\multirow{ 2}{*}{0.0949}	&	\multirow{ 2}{*}{0.0382}	&	\multirow{ 2}{*}{0.6910}	&	\multirow{ 2}{*}{0.8179}	&	\multirow{ 2}{*}{0.4215}	&	\multirow{ 2}{*}{0.6669}	&	\multirow{ 2}{*}{0.6910}	&	\multirow{ 2}{*}{0.8179}	&	\multirow{ 2}{*}{0.7235}	&	\multirow{ 2}{*}{0.8118}	\\
				$\beta=1$	&		&		&		&		&		&		&		&		&		&		\\
				FSMLP\textsubscript{struct},	&	\multirow{ 2}{*}{0.0932}	&	\multirow{ 2}{*}{0.0346}	&	\multirow{ 2}{*}{0.6921}	&	\multirow{ 2}{*}{0.8571}	&	\multirow{ 2}{*}{0.4171}	&	\multirow{ 2}{*}{0.7029}	&	\multirow{ 2}{*}{0.6921}	&	\multirow{ 2}{*}{0.8571}	&	\multirow{ 2}{*}{0.7412}	&	\multirow{ 2}{*}{0.8588}	\\
				$\beta=10$	&		&		&		&		&		&		&		&		&		&		\\
				\hline
		\end{tabular}}
	\end{table}
	As we have already discussed in Sec. \ref{sec:chap4_proposed} the lesser the value of SS, the better the projected space (formed by selected features) preserves the original pairwise distances and hence the structure of the original data. We observe in Table \ref{tab:perf_ecoli_train}, the mutual information based method shows the lowest value of SS, and the second lowest is FSMLP\textsubscript{struct} with $\beta=10$ for both $Q=3$ and $Q=4$. Actually, the SS values for the mutual information based method and FSMLP\textsubscript{struct} with $\beta=10$ are almost the same, equal up to two places after decimal points in both choices of $Q$. The SS values achieved by ICA, the F score based method and  FSMLP are comparatively higher. So,  the mutual information based method and FSMLP\textsubscript{struct} with $\beta=10$ preserve the original pairwise distances most in  the projected space. They are also expected to preserve the structures most. The values of the other three structure preserving measures i.e., NMI, ARI, and JI confirm that. We know that the higher the values of NMI, ARI, and JI are, the closer the clustering structures of the projected space are to the original clustering structures. The highest values of the NMI, ARI, and JI are obtained by the mutual information based method, followed by the FSMLP\textsubscript{struct} with $\beta=10$. So, in cluster structure preservation, mutual information based method and  FSMLP\textsubscript{struct} with $\beta=10$ perform better than the other three methods and even than the other two models trained by FSMLP\textsubscript{struct} with $\beta=0.1$ and $1$. $\beta$ is the weight of the regualizer $E_{sammons}$ in (\ref{eq:sammons}). Although SS and $E_{sammons}$ are not exactly same, under the influence of $E_{select}$, it is expected that the higher the value of $\beta$ lesser the value of SS would be. Table \ref{tab:perf_ecoli_train} reconfirms that. The SS values become lesser as the $\beta$ increases from $0.1$ to $10$. SS values of FSMLP (which is basically FSMLP\textsubscript{struct} with $\beta=0$) and  FSMLP\textsubscript{struct} with $\beta=0.1$ are the same for both the choices of $Q$. Actually, FSMLP  and  FSMLP\textsubscript{struct} with $\beta=0.1$ have all the ten measures the same. It proves that for the E.coli dataset $\beta=0.1$ does not give any effective weightage to the structure preservation term and chooses the same subsets as FSMLP. For the classification performance measure OCA, FSMLP\textsubscript{struct} with $\beta=1$, achieves the highest value, followed by FSMLP\textsubscript{struct} with $\beta=10$. The mutual information based method and FSMLP\textsubscript{struct} with $\beta=10$ have all the structure preserving measures either almost equal or of comparable values, however for OCA, FSMLP\textsubscript{struct} with $\beta=10$ is better than mutual information based method with a margin more than $18\%$. For E. coli data, the test set follows the observed trends in the training set with the following exceptions. First, for $Q=3$, the values of NMI, ARI, and JI have not increased as $\beta$ increases from $0.1$ to $10$. Second, for $Q=4$, FSMLP\textsubscript{struct} with $\beta=10$ beats all the methods including mutual information based method. Analyzing the performances over train and test sets, for E. coli data FSMLP\textsubscript{struct} with $\beta=10$ is the winner among the other six models.
	
	Tables \ref{tab:perf_glass_train} and \ref{tab:perf_glass_test} compare the performance of the proposed method with other methods in terms of different criteria for the Glass dataset on its training and test sets, respectively. 
	
	\begin{table}[!h]
		\caption{Performance comparison for Glass, training set for different choices of $\beta$ and $Q$.}\label{tab:perf_glass_train}
		\centering
		\resizebox{12.5cm}{!}{
			\begin{tabular}{|c|c c c c c c c c c c|}
				\hline
				Method	&	\multicolumn{2}{c}{SS}			&	\multicolumn{2}{c}{NMI}			&\multicolumn{2}{c}{ARI}			&	\multicolumn{2}{c}{JI}			&	\multicolumn{2}{c|}{OCA}			\\
				&	Q=4	&	Q=5	&	Q=4	&	Q=5	&	Q=4	&	Q=5	&	Q=4	&	Q=5	&	Q=4	&	Q=5	\\
				\hline
				ICA	&	0.2699	&	0.2216	&	0.5066	&	0.5416	&	0.3343	&	0.3688	&	0.5066	&	0.5416	&	0.6771	&	0.6823	\\
				F Score	&	0.1284	&	0.1011	&	0.5961	&	0.6343	&	0.5137	&	0.5750	&	0.5961	&	0.6343	&	0.6979	&	0.7396	\\
				Mutual Info	&	0.2091	&	0.1159	&	0.5073	&	0.6113	&	0.3321	&	0.5688	&	0.5073	&	0.6113	&	0.6823	&	0.6719	\\
				FSMLP	&	0.1968	&	0.1593	&	0.5512	&	0.6028	&	0.4473	&	0.5235	&	0.5512	&	0.6028	&	0.6302	&	0.7094	\\
				FSMLP\textsubscript{struct},	&	\multirow{ 2}{*}{0.1968}	&	\multirow{ 2}{*}{0.0747}	&	\multirow{ 2}{*}{0.5512}	&	\multirow{ 2}{*}{0.7501}	&	\multirow{ 2}{*}{0.4473}	&	\multirow{ 2}{*}{0.6873}	&	\multirow{ 2}{*}{0.5512}	&	\multirow{ 2}{*}{0.7501}	&	\multirow{ 2}{*}{0.6302}	&	\multirow{ 2}{*}{0.7510}	\\
				$\beta=0.1$	&		&		&		&		&		&		&		&		&		&		\\
				FSMLP\textsubscript{struct},	&	\multirow{ 2}{*}{0.0494}	&	\multirow{ 2}{*}{0.0623}	&	\multirow{ 2}{*}{0.7732}	&	\multirow{ 2}{*}{0.7965}	&	\multirow{ 2}{*}{0.7124}	&	\multirow{ 2}{*}{0.7665}	&	\multirow{ 2}{*}{0.7732}	&	\multirow{ 2}{*}{0.7965}	&	\multirow{ 2}{*}{0.7740}	&	\multirow{ 2}{*}{0.7708}	\\
				$\beta=1$	&		&		&		&		&		&		&		&		&		&		\\
				FSMLP\textsubscript{struct},	&	\multirow{ 2}{*}{0.0309}	&	\multirow{ 2}{*}{0.0406}	&	\multirow{ 2}{*}{0.8551}	&	\multirow{ 2}{*}{0.7394}	&	\multirow{ 2}{*}{0.8507}	&	\multirow{ 2}{*}{0.6792}	&	\multirow{ 2}{*}{0.8551}	&	\multirow{ 2}{*}{0.7394}	&	\multirow{ 2}{*}{0.8125}	&	\multirow{ 2}{*}{0.7667}	\\
				$\beta=10$	&		&		&		&		&		&		&		&		&		&		\\
				\hline
		\end{tabular}}
	\end{table}
	
	\begin{table}[!h]
		\caption{Performance comparison for Glass, test set for different choices of $\beta$ and $Q$.}\label{tab:perf_glass_test}
		\centering
		\resizebox{12.5cm}{!}{
			\begin{tabular}{|c|c c c c c c c c c c|}
				\hline
				Method	&	\multicolumn{2}{c}{SS}			&	\multicolumn{2}{c}{NMI}			&\multicolumn{2}{c}{ARI}			&	\multicolumn{2}{c}{JI}			&	\multicolumn{2}{c|}{OCA}			\\
				&	Q=4	&	Q=5	&	Q=4	&	Q=5	&	Q=4	&	Q=5	&	Q=4	&	Q=5	&	Q=4	&	Q=5	\\
				\hline
				ICA	&	0.3039	&	0.2333	&	0.6536	&	0.6063	&	0.3409	&	0.2979	&	0.6536	&	0.6063	&	0.4545	&	0.4545	\\
				F Score	&	0.1393	&	0.1258	&	0.7028	&	0.7744	&	0.5746	&	0.6861	&	0.7028	&	0.7744	&	0.5455	&	0.6364	\\
				Mutual Info	&	0.2034	&	0.1372	&	0.7805	&	0.7728	&	0.6358	&	0.6655	&	0.7805	&	0.7728	&	0.4545	&	0.5909	\\
				FSMLP	&	0.2027	&	0.1669	&	0.7366	&	0.7425	&	0.5053	&	0.5380	&	0.7366	&	0.7425	&	0.5727	&	0.6727	\\
				FSMLP\textsubscript{struct},	&	\multirow{ 2}{*}{0.2027}	&	\multirow{ 2}{*}{0.0806}	&	\multirow{ 2}{*}{0.7366}	&	\multirow{ 2}{*}{0.7885}	&	\multirow{ 2}{*}{0.5053}	&	\multirow{ 2}{*}{0.6190}	&	\multirow{ 2}{*}{0.7366}	&	\multirow{ 2}{*}{0.7885}	&	\multirow{ 2}{*}{0.5727}	&	\multirow{ 2}{*}{0.6636}	\\
				$\beta=0.1$	&		&		&		&		&		&		&		&		&		&		\\
				FSMLP\textsubscript{struct},	&	\multirow{ 2}{*}{0.0406}	&	\multirow{ 2}{*}{0.0636}	&	\multirow{ 2}{*}{0.8265}	&	\multirow{ 2}{*}{0.8300}	&	\multirow{ 2}{*}{0.6951}	&	\multirow{ 2}{*}{0.7355}	&	\multirow{ 2}{*}{0.8265}	&	\multirow{ 2}{*}{0.8300}	&	\multirow{ 2}{*}{0.6818}	&	\multirow{ 2}{*}{0.6727}	\\
				$\beta=1$	&		&		&		&		&		&		&		&		&		&		\\
				FSMLP\textsubscript{struct},	&	\multirow{ 2}{*}{0.0215}	&	\multirow{ 2}{*}{0.0372}	&	\multirow{ 2}{*}{0.9022}	&	\multirow{ 2}{*}{0.7950}	&	\multirow{ 2}{*}{0.8670}	&	\multirow{ 2}{*}{0.6210}	&	\multirow{ 2}{*}{0.9022}	&	\multirow{ 2}{*}{0.7950}	&	\multirow{ 2}{*}{0.6727}	&	\multirow{ 2}{*}{0.6818}	\\
				$\beta=10$	&		&		&		&		&		&		&		&		&		&		\\
				
				\hline
		\end{tabular}}
	\end{table}
	The chosen numbers of features for the Glass data are $4$ and $5$. The expected nature of decreasing SS with increasing $\beta$ is clearly observed for $Q=5$ for both the training and test set. For $Q=4$, the Glass data also follows the characteristics of the E. coli data of having the same values for FSMLP and FSMLP\textsubscript{struct} with $\beta=0.1$ in all the ten measures for both training and test set.  For $Q=4$, from $\beta=0.1$ onwards, increasing $\beta$s produce decreasing SS values and  increasing NMI, ARI, and JI values for both training and test datasets. We observe from the Tables \ref{tab:perf_glass_train} and \ref{tab:perf_glass_test}, for $Q=5$, as the $\beta$ increases from $0$ (FSMLP) to $0.1$, and then to $1$, NMI, ARI, and JI values are increased for both training and test datasets, however at $\beta=10$,  NMI, ARI, and JI values are decreased compared to $\beta=0.1$ and $1$. We can conclude that, for $Q=4$, FSMLP\textsubscript{struct} with $\beta=10$ gives the best structure preserving performance among the considered models and for $Q=5$, FSMLP\textsubscript{struct} with $\beta=1$ is best in structure preservation. In terms of the classification performance measure OCA, FSMLP\textsubscript{struct} with $\beta=10$ and FSMLP\textsubscript{struct} with $\beta=1 $ show the highest OCA values for the training set  and test set respectively, with $Q=4$. On the other hand, for $Q=5$, FSMLP\textsubscript{struct} with $\beta=10$ show the highest OCA values for the training set and FSMLP\textsubscript{struct} with $\beta=1$ show the highest OCA values for the test set. Inspecting all the performance measure values, we conclude that for the Glass dataset, both FSMLP\textsubscript{struct} with $\beta=10$ and FSMLP\textsubscript{struct} with $\beta=1$ are comparatively better in simultaneously preserving both class and cluster structures than the other methods.
	
	The performances of the Ionosphere dataset are recorded in Tables \ref{tab:perf_ionosphere_train} and \ref{tab:perf_ionosphere_test} for training and test sets respectively. 
	\begin{table}[!h]
		\caption{Performance comparison for Ionosphere, training set for different choices of $\beta$ and $Q$.}\label{tab:perf_ionosphere_train}
		\centering
		\resizebox{12.5cm}{!}{
			\begin{tabular}{|c|c c c c c c c c c c|}
				\hline
				Method	&	\multicolumn{2}{c}{SS}			&	\multicolumn{2}{c}{NMI}			&\multicolumn{2}{c}{ARI}			&	\multicolumn{2}{c}{JI}			&	\multicolumn{2}{c|}{OCA}			\\
				&	Q=12	&	Q=17	&	Q=12	&	Q=17	&	Q=12	&	Q=17	&	Q=12	&	Q=17	&	Q=12	&	Q=17\\
				\hline
				ICA	&	0.1791	&	0.0926	&	0.8261	&	0.8817	&	0.8766	&	0.9250	&	0.8261	&	0.8817	&	0.9492	&	0.9714	\\
				F Score	&	0.1753	&	0.0914	&	0.9035	&	0.9313	&	0.9497	&	0.9621	&	0.9035	&	0.9313	&	0.9397	&	0.9746	\\
				Mutual Info	&	0.1776	&	0.0803	&	0.6829	&	0.8645	&	0.7838	&	0.9250	&	0.6829	&	0.8645	&	0.9746	&	0.9651	\\
				FSMLP	&	0.1721	&	0.0895	&	0.6292	&	0.7330	&	0.7246	&	0.8192	&	0.6292	&	0.7330	&	0.9606	&	0.9733	\\
				FSMLP\textsubscript{struct},	&	\multirow{ 2}{*}{0.1675}	&	\multirow{ 2}{*}{0.0878}	&	\multirow{ 2}{*}{0.6689}	&	\multirow{ 2}{*}{0.7415}	&	\multirow{ 2}{*}{0.7641}	&	\multirow{ 2}{*}{0.8243}	&	\multirow{ 2}{*}{0.6689}	&	\multirow{ 2}{*}{0.7415}	&	\multirow{ 2}{*}{0.9638}	&	\multirow{ 2}{*}{0.9733}	\\
				$\beta=0.1$	&																&		&		\\
				FSMLP\textsubscript{struct},	&	\multirow{ 2}{*}{0.1505}	&	\multirow{ 2}{*}{0.0776}	&	\multirow{ 2}{*}{0.7307}	&	\multirow{ 2}{*}{0.7867}	&	\multirow{ 2}{*}{0.8051}	&	\multirow{ 2}{*}{0.8582}	&	\multirow{ 2}{*}{0.7307}	&	\multirow{ 2}{*}{0.7867}	&	\multirow{ 2}{*}{0.9632}	&	\multirow{ 2}{*}{0.9568}	\\
				$\beta=1$	&																&		&		\\
				FSMLP\textsubscript{struct},	&	\multirow{ 2}{*}{0.1437}	&	\multirow{ 2}{*}{0.0766}	&	\multirow{ 2}{*}{0.7894}	&	\multirow{ 2}{*}{0.8111}	&	\multirow{ 2}{*}{0.8533}	&	\multirow{ 2}{*}{0.8723}	&	\multirow{ 2}{*}{0.7894}	&	\multirow{ 2}{*}{0.8111}	&	\multirow{ 2}{*}{0.9683}	&	\multirow{ 2}{*}{0.9644}	\\
				$\beta=10$	&		&		&		&		&		&		&		&		&		&		\\
				\hline
		\end{tabular}}
	\end{table}
	
	\begin{table}[!h]
		\caption{Performance comparison for Ionosphere, test set for different choices of $\beta$ and $Q$.}\label{tab:perf_ionosphere_test}
		\centering
		\resizebox{12.5cm}{!}{
			\begin{tabular}{|c|c c c c c c c c c c|}
				\hline
				Method	&	\multicolumn{2}{c}{SS}			&	\multicolumn{2}{c}{NMI}			&\multicolumn{2}{c}{ARI}			&	\multicolumn{2}{c}{JI}			&	\multicolumn{2}{c|}{OCA}			\\
				&	Q=12	&	Q=17	&	Q=12	&	Q=17	&	Q=12	&	Q=17	&	Q=12	&	Q=17	&	Q=12	&	Q=17\\
				\hline
				ICA	&	0.1652	&	0.0964	&	0.7351	&	0.7351	&	0.7782	&	0.7782	&	0.7351	&	0.7351	&	0.9143	&	0.9429	\\
				F Score	&	0.1952	&	0.1003	&	0.4325	&	0.4778	&	0.3411	&	0.4152	&	0.4325	&	0.4778	&	0.9429	&	0.9429	\\
				Mutual Info	&	0.2186	&	0.0972	&	0.5287	&	0.6541	&	0.4959	&	0.6774	&	0.5287	&	0.6541	&	0.9429	&	0.9429	\\
				FSMLP	&	0.1801	&	0.0913	&	0.6518	&	0.8284	&	0.6841	&	0.8682	&	0.6518	&	0.8284	&	0.9371	&	0.9143	\\
				FSMLP\textsubscript{struct},	&	\multirow{ 2}{*}{0.1722}	&	\multirow{ 2}{*}{0.0885}	&	\multirow{ 2}{*}{0.7434}	&	\multirow{ 2}{*}{0.8284}	&	\multirow{ 2}{*}{0.7714}	&	\multirow{ 2}{*}{0.8682}	&	\multirow{ 2}{*}{0.7434}	&	\multirow{ 2}{*}{0.8284}	&	\multirow{ 2}{*}{0.9371}	&	\multirow{ 2}{*}{0.9314}	\\
				$\beta=0.1$	&		&		&		&		&		&		&		&		&		&		\\
				FSMLP\textsubscript{struct},	&	\multirow{ 2}{*}{0.1437}	&	\multirow{ 2}{*}{0.0726}	&	\multirow{ 2}{*}{0.7764}	&	\multirow{ 2}{*}{0.8597}	&	\multirow{ 2}{*}{0.8265}	&	\multirow{ 2}{*}{0.8884}	&	\multirow{ 2}{*}{0.7764}	&	\multirow{ 2}{*}{0.8597}	&	\multirow{ 2}{*}{0.8971}	&	\multirow{ 2}{*}{0.9257}	\\
				$\beta=1$	&		&		&		&		&		&		&		&		&		&		\\
				FSMLP\textsubscript{struct},	&	\multirow{ 2}{*}{0.1310}	&	\multirow{ 2}{*}{0.0720}	&	\multirow{ 2}{*}{0.7960}	&	\multirow{ 2}{*}{0.9146}	&	\multirow{ 2}{*}{0.8454}	&	\multirow{ 2}{*}{0.9328}	&	\multirow{ 2}{*}{0.7960}	&	\multirow{ 2}{*}{0.9146}	&	\multirow{ 2}{*}{0.9029}	&	\multirow{ 2}{*}{0.9257}	\\
				$\beta=10$	&		&		&		&		&		&		&		&		&		&		\\
				
				\hline
		\end{tabular}}
	\end{table}
	For the Ionosphere data set, the number of selected features, $Q$ is set as $12$ and $17$. Here, in all the cases, whenever the $\beta$ is increasing, SS is decreasing  and the other structure preserving indices NMI, ARI, and JI are increasing consistently. Unlike, E. coli and Glass data set, here when $\beta$ increases from $0$ (in FSMLP) to $0.1$, the structure preserving metrics including SS shifted in the desired direction in most of the cases and remained the same in some cases. Except for SS, in the other three structure preserving measures, ICA and F score based method have performed better than FSMLP and FSMLP\textsubscript{struct} for all the cases. Classification performance is good for almost all methods for the Ionosphere data set. In the training set, for both $Q=12$ and $Q=17$, an accuracy of $97.46\%$ is reached by mutual info and F score based methods, however, FSMLP and FSMLP\textsubscript{struct} models have reached more than $96\%$ accuracy in every case. For the test set, all the structure preserving indices are better for FSMLP and FSMLP\textsubscript{struct} than ICA, F score, and mutual information based methods, although in terms of classification score OCA, the F score and mutual information based methods have performed marginally better than FSMLP and FSMLP\textsubscript{struct} models. This may have happened because the selected features from the neural network based classifier which are expected to be discriminatory features, may not be the best for SVM. Moreover, FSMLP\textsubscript{struct} makes a compromise between preserving cluster structure and classifier loss. For the Ionosphere dataset, our proposed models are not the winner. May be with higher $\beta$, FSMLP\textsubscript{struct} would deliver better scores.
	
	For the Sonar data, the summary of the performances of the training and test data sets  in terms of the five measures for two choices of the number of selected features are  available in Tables  \ref{tab:perf_sonar_train} and \ref{tab:perf_sonar_test}.
	\begin{table}[!tb]
		\caption{Performance comparison for Sonar, training set for different choices of $\beta$ and $Q$.}\label{tab:perf_sonar_train}
		\centering
		\resizebox{12.5cm}{!}{
			\begin{tabular}{|c|c c c c c c c c c c|}
				\hline
				Method	&	\multicolumn{2}{c}{SS}			&	\multicolumn{2}{c}{NMI}			&\multicolumn{2}{c}{ARI}			&	\multicolumn{2}{c}{JI}			&	\multicolumn{2}{c|}{OCA}			\\
				&	Q=21	&	Q=30	&	Q=21	&	Q=30	&	Q=21	&	Q=30	&	Q=21	&	Q=30	&	Q=21	&	Q=30	\\
				\hline
				ICA	&	0.7830	&	0.4963	&	0.0004	&	0.0276	&	-0.0040	&	0.0382	&	0.0004	&	0.0276	&	0.7112	&	0.8075	\\
				F Score	&	0.3614	&	0.2029	&	0.1167	&	0.1778	&	0.1159	&	0.1926	&	0.1167	&	0.1778	&	0.8289	&	0.9733	\\
				Mutual Info	&	0.2247	&	0.0950	&	0.4601	&	0.6601	&	0.5662	&	0.7585	&	0.4601	&	0.6601	&	0.9733	&	0.9465	\\
				FSMLP	&	0.1880	&	0.0910	&	0.3735	&	0.5293	&	0.4432	&	0.6104	&	0.3735	&	0.5293	&	0.9273	&	0.9765	\\
				FSMLP\textsubscript{struct},	&	\multirow{ 2}{*}{0.1314}	&	\multirow{ 2}{*}{0.0779}	&	\multirow{ 2}{*}{0.4917}	&	\multirow{ 2}{*}{0.5587}	&	\multirow{ 2}{*}{0.5698}	&	\multirow{ 2}{*}{0.6358}	&	\multirow{ 2}{*}{0.4917}	&	\multirow{ 2}{*}{0.5587}	&	\multirow{ 2}{*}{0.9775}	&	\multirow{ 2}{*}{0.9722}	\\
				$\beta=0.1$	&		&		&		&		&		&		&		&		&		&		\\
				FSMLP\textsubscript{struct},	&	\multirow{ 2}{*}{0.0620}	&	\multirow{ 2}{*}{0.0285}	&	\multirow{ 2}{*}{0.5743}	&	\multirow{ 2}{*}{0.6838}	&	\multirow{ 2}{*}{0.6620}	&	\multirow{ 2}{*}{0.7696}	&	\multirow{ 2}{*}{0.5743}	&	\multirow{ 2}{*}{0.6838}	&	\multirow{ 2}{*}{0.9872}	&	\multirow{ 2}{*}{0.9840}	\\
				$\beta=1$	&		&		&		&		&		&		&		&		&		&		\\
				FSMLP\textsubscript{struct},	&	\multirow{ 2}{*}{0.0481}	&	\multirow{ 2}{*}{0.0201}	&	\multirow{ 2}{*}{0.6983}	&	\multirow{ 2}{*}{0.7473}	&	\multirow{ 2}{*}{0.7751}	&	\multirow{ 2}{*}{0.8318}	&	\multirow{ 2}{*}{0.6983}	&	\multirow{ 2}{*}{0.7473}	&	\multirow{ 2}{*}{0.9936}	&	\multirow{ 2}{*}{0.9968}	\\
				$\beta=10$	&		&		&		&		&		&		&		&		&		&		\\
				\hline
		\end{tabular}}
	\end{table}
	\begin{table}[!tb]
		\caption{Performance comparison for Sonar, test set for different choices of $\beta$ and $Q$.}\label{tab:perf_sonar_test}
		\centering
		\resizebox{12.5cm}{!}{
			\begin{tabular}{|c|c c c c c c c c c c|}
				\hline
				Method	&	\multicolumn{2}{c}{SS}			&	\multicolumn{2}{c}{NMI}			&\multicolumn{2}{c}{ARI}			&	\multicolumn{2}{c}{JI}			&	\multicolumn{2}{c|}{OCA}			\\
				&	Q=21	&	Q=30	&	Q=21	&	Q=30	&	Q=21	&	Q=30	&	Q=21	&	Q=30	&	Q=21	&	Q=30	\\
				\hline
				ICA	&	0.7556	&	0.4968	&	0.2513	&	0.1993	&	0.1546	&	0.2383	&	0.2513	&	0.1993	&	0.7143	&	0.5714	\\
				F Score	&	0.3629	&	0.1941	&	0.2906	&	0.3029	&	0.3519	&	0.3536	&	0.2906	&	0.3029	&	0.5714	&	0.8571	\\
				Mutual Info	&	0.2144	&	0.0955	&	0.2906	&	0.5411	&	0.3519	&	0.6378	&	0.2906	&	0.5411	&	0.8095	&	0.8095	\\
				FSMLP	&	0.1967	&	0.0977	&	0.3918	&	0.5288	&	0.4633	&	0.5850	&	0.3918	&	0.5288	&	0.8190	&	0.8857	\\
				FSMLP\textsubscript{struct},	&	\multirow{ 2}{*}{0.1436}	&	\multirow{ 2}{*}{0.0840}	&	\multirow{ 2}{*}{0.4026}	&	\multirow{ 2}{*}{0.5540}	&	\multirow{ 2}{*}{0.4368}	&	\multirow{ 2}{*}{0.6155}	&	\multirow{ 2}{*}{0.4026}	&	\multirow{ 2}{*}{0.5540}	&	\multirow{ 2}{*}{0.7714}	&	\multirow{ 2}{*}{0.8762}	\\
				$\beta=0.1$	&		&		&		&		&		&		&		&		&		&		\\
				FSMLP\textsubscript{struct},	&	\multirow{ 2}{*}{0.0679}	&	\multirow{ 2}{*}{0.0302}	&	\multirow{ 2}{*}{0.6905}	&	\multirow{ 2}{*}{0.6824}	&	\multirow{ 2}{*}{0.7223}	&	\multirow{ 2}{*}{0.7146}	&	\multirow{ 2}{*}{0.6905}	&	\multirow{ 2}{*}{0.6824}	&	\multirow{ 2}{*}{0.7619}	&	\multirow{ 2}{*}{0.8381}	\\
				$\beta=1$	&		&		&		&		&		&		&		&		&		&		\\
				FSMLP\textsubscript{struct},	&	\multirow{ 2}{*}{0.0473}	&	\multirow{ 2}{*}{0.0198}	&	\multirow{ 2}{*}{0.5001}	&	\multirow{ 2}{*}{0.7156}	&	\multirow{ 2}{*}{0.5547}	&	\multirow{ 2}{*}{0.7490}	&	\multirow{ 2}{*}{0.5001}	&	\multirow{ 2}{*}{0.7156}	&	\multirow{ 2}{*}{0.8000}	&	\multirow{ 2}{*}{0.7810}	\\
				$\beta=10$	&		&		&		&		&		&		&		&		&		&		\\
				\hline
		\end{tabular}}
	\end{table}
	We set, $Q=21$ and $30$ for the Sonar data set. In the case of the Sonar data set, not only with increasing $\beta $, all the structure preserving indices improve, in case of the training set, FSMLP\textsubscript{struct} with $\beta=10$ are significantly better than ICA, F score, and mutual information based methods, and FSMLP in all five scores for both the choices of $Q$. In test set for some cases, FSMLP\textsubscript{struct} with $\beta=1$ is better than FSMLP\textsubscript{struct} with $\beta=10$. For the Sonar data set, clearly, the proposed method performed extremely well in terms of classification and clustering performance. 
	
	Tables \ref{tab:perf_AR_train} and \ref{tab:perf_AR_test} summarize the performances of the proposed method and other comparing methods for training and test sets, respectively for the AR10P data set. The original number of features, $P$ for the AR10P data set is $2400$, which is comparatively higher than that of the other two data sets used in this sub-section. The two choices of the number of selected features here are $40$ and $60$ and these are not approximately $35\%$ and $50\%$  of the original dimension like in previous cases. 
	\begin{table}[!tb]
		\caption{Performance comparison for AR10P, training set for different choices of $\beta$ and $Q$.}\label{tab:perf_AR_train}
		\centering
		\resizebox{12.5cm}{!}{
			\begin{tabular}{|c|c c c c c c c c c c|}
				\hline
				Method	&	\multicolumn{2}{c}{SS}			&	\multicolumn{2}{c}{NMI}			&\multicolumn{2}{c}{ARI}			&	\multicolumn{2}{c}{JI}			&	\multicolumn{2}{c|}{OCA}			\\
				
				&	Q=40	&	Q=60	&	Q=40	&	Q=60	&	Q=40	&	Q=60	&	Q=40	&	Q=60	&	Q=40	&	Q=60	\\
				\hline
				ICA	&	0.7619	&	0.7107	&	0.2807	&	0.2449	&	0.2222	&	0.2066	&	0.2807	&	0.2449	&	1	&	1	\\
				F Score	&	0.7933	&	0.7484	&	0.1693	&	0.1709	&	0.0550	&	0.0438	&	0.1693	&	0.1709	&	1	&	1	\\
				Mutual Info	&	0.7805	&	0.7309	&	0.1613	&	0.1863	&	0.0429	&	0.0558	&	0.1613	&	0.1863	&	0.9915	&	0.9915	\\
				FSMLP	&	0.7640	&	0.7109	&	0.2296	&	0.2108	&	0.1038	&	0.0943	&	0.2296	&	0.2108	&	0.9983	&	1	\\
				FSMLP\textsubscript{struct},	&	\multirow{ 2}{*}{0.7566}	&	\multirow{ 2}{*}{0.7061}	&	\multirow{ 2}{*}{0.3828}	&	\multirow{ 2}{*}{0.5251}	&	\multirow{ 2}{*}{0.3907}	&	\multirow{ 2}{*}{0.5654}	&	\multirow{ 2}{*}{0.3828}	&	\multirow{ 2}{*}{0.5251}	&	\multirow{ 2}{*}{1}	&	\multirow{ 2}{*}{1}	\\
				$\beta=25$	&		&		&		&		&		&		&		&		&		&		\\
				FSMLP\textsubscript{struct},	&	\multirow{ 2}{*}{0.7567}	&	\multirow{ 2}{*}{0.7059}	&	\multirow{ 2}{*}{0.5307}	&	\multirow{ 2}{*}{0.5089}	&	\multirow{ 2}{*}{0.5575}	&	\multirow{ 2}{*}{0.5268}	&	\multirow{ 2}{*}{0.5307}	&	\multirow{ 2}{*}{0.5089}	&	\multirow{ 2}{*}{1}	&	\multirow{ 2}{*}{1}	\\
				$\beta=50$	&		&		&		&		&		&		&		&		&		&		\\
				FSMLP\textsubscript{struct},	&	\multirow{ 2}{*}{0.7564}	&	\multirow{ 2}{*}{0.7061}	&	\multirow{ 2}{*}{0.4849}	&	\multirow{ 2}{*}{0.5830}	&	\multirow{ 2}{*}{0.4852}	&	\multirow{ 2}{*}{0.6353}	&	\multirow{ 2}{*}{0.4849}	&	\multirow{ 2}{*}{0.5830}	&	\multirow{ 2}{*}{1}	&	\multirow{ 2}{*}{1}	\\
				$\beta=100$	&		&		&		&		&		&		&		&		&		&		\\
				
				\hline
		\end{tabular}}
	\end{table}
	\begin{table}[!tb]
		\caption{Performance comparison for AR10P, test set for different choices of $\beta$ and $Q$.}\label{tab:perf_AR_test}
		\centering
		\resizebox{12.5cm}{!}{
			\begin{tabular}{|c|c c c c c c c c c c|}
				\hline
				Method	&	\multicolumn{2}{c}{SS}			&	\multicolumn{2}{c}{NMI}			&\multicolumn{2}{c}{ARI}			&	\multicolumn{2}{c}{JI}			&	\multicolumn{2}{c|}{OCA}			\\
				
				&	Q=40	&	Q=60	&	Q=40	&	Q=60	&	Q=40	&	Q=60	&	Q=40	&	Q=60	&	Q=40	&	Q=60	\\
				\hline
				ICA	&	0.7739	&	0.7256	&	0.8255	&	0.8222	&	0.0031	&	-0.0523	&	0.8255	&	0.8222	&	0.6154	&	0.6154	\\
				F Score	&	0.7364	&	0.6856	&	0.9072	&	0.9152	&	0.3645	&	0.4155	&	0.9072	&	0.9152	&	0.8462	&	0.8462	\\
				Mutual Info	&	0.7614	&	0.7112	&	0.8451	&	0.8451	&	0.0122	&	0.0122	&	0.8451	&	0.8451	&	0.9231	&	0.8462	\\
				FSMLP	&	0.7838	&	0.7213	&	0.9051	&	0.9152	&	0.3686	&	0.4155	&	0.9051	&	0.9152	&	0.8462	&	0.7538	\\
				FSMLP\textsubscript{struct},	&	\multirow{ 2}{*}{0.7795}	&	\multirow{ 2}{*}{0.7346}	&	\multirow{ 2}{*}{0.9223}	&	\multirow{ 2}{*}{0.9146}	&	\multirow{ 2}{*}{0.4820}	&	\multirow{ 2}{*}{0.4380}	&	\multirow{ 2}{*}{0.9223}	&	\multirow{ 2}{*}{0.9146}	&	\multirow{ 2}{*}{0.4615}	&	\multirow{ 2}{*}{0.6154}	\\
				$\beta=25$	&		&		&		&		&		&		&		&		&		&		\\
				FSMLP\textsubscript{struct},	&	\multirow{ 2}{*}{0.7815}	&	\multirow{ 2}{*}{0.7352}	&	\multirow{ 2}{*}{0.9050}	&	\multirow{ 2}{*}{0.9050}	&	\multirow{ 2}{*}{0.3782}	&	\multirow{ 2}{*}{0.3782}	&	\multirow{ 2}{*}{0.9050}	&	\multirow{ 2}{*}{0.9050}	&	\multirow{ 2}{*}{0.4615}	&	\multirow{ 2}{*}{0.6154}	\\
				$\beta=50$	&		&		&		&		&		&		&		&		&		&		\\
				FSMLP\textsubscript{struct},	&	\multirow{ 2}{*}{0.7804}	&	\multirow{ 2}{*}{0.7356}	&	\multirow{ 2}{*}{0.9056}	&	\multirow{ 2}{*}{0.9136}	&	\multirow{ 2}{*}{0.3557}	&	\multirow{ 2}{*}{0.4727}	&	\multirow{ 2}{*}{0.9056}	&	\multirow{ 2}{*}{0.9136}	&	\multirow{ 2}{*}{0.4615}	&	\multirow{ 2}{*}{0.6154}	\\
				$\beta=100$	&		&		&		&		&		&		&		&		&		&		\\
				
				\hline
		\end{tabular}}
	\end{table}
	The study in \cite{chakraborty2014feature}, proposed a feature selection scheme for redundancy control in features. They reported an average number of selected features of $58.9$ without practicing redundancy control and an average number of selected features in the range of $22.8$ to $44.2$ when practicing redundancy control for AR10P data set. Hence, we choose the number of selected features $Q$ as $40$ and $60$. From the classification scores shown in Table \ref{tab:perf_AR_train}, we note that for all the  methods for both the choices of $Q$, classification scores in training set are more than $99\%$. In the training set, we observe that for FSMLP\textsubscript{struct} as $\beta$ increases SS is decreased in almost all the cases. But for the test set, this is not true. For the other structure-preserving measures for the training set, FSMLP\textsubscript{struct} with $\beta=50$ is best among all the methods for $Q=40$  and FSMLP\textsubscript{struct} with $\beta=100$ is best among all the methods for $Q=60$. In the test set, all the methods have performed almost the same in terms of the structure-preserving measures. The classification performances of FSMLP\textsubscript{struct} are very poor in the test set for AR10P data. The significant differences in training and test OCA values for FSMLP\textsubscript{struct} indicate poor generalization of the system. This problem may be addressed by choosing the number of nodes for our MLP based model through cross-validation.
	
	 Results from the five data sets clearly establish the benefit of introducing the proposed structure preserving regularizer term, $E_{sammons}$ in the overall loss function (\ref{eq:overall_error_nonlarge_n}) of the MLP based embedded feature selection scheme. Next we shall consider the band (channel) selection problem for hyperspectral satelite images.
	
	\subsection{Band selection in hyperspectral images}\label{subsec:chap4_expt_HSI}
	
	Let our considered hyperspectral image $I$ be of dimension, $H\times W\times P$ where, $H$, $W$, and $P$ are the height, width, and number of spectral bands of the image respectively. We can represent the pixels of $I$ as $\mathbf{x}_{i}\in \mathbb{R}^{P}: i=1, 2, \dots ,H\times W$. Let, there be total $n$ pixels annotated with $C$ land cover classes. Without any loss of generality, we take the first $n$ pixels, i.e., $i=1,2, \dots n$ as the pixels having class labels. Our input data for land cover classification problem be $\mathbf{X}=\{\mathbf{x}_{i}=(x_{i1},x_{i2},\cdots,x_{iP}) \in \mathbb{R}^{P}\}_{i=1}^{n}$. The collection of class labels of  $\mathbf{X}$ be $\mathbf{Z}=\{z_{i} \in\{1,2,\cdots, C\}\}_{i=1}^{n}$, where, $z_{i}$ is the class label corresponding to $\mathbf{x}_{i}$. We aim to select a subset of size $Q$ from the original set of bands such that the selected subset performs reasonably well for land cover classification as well as in clustering. 
	We have performed the experiments with three benchmark HSI datasets for land cover classification problems- Indian pines, Pavia University, and Salinas\cite{hyperspectral2021}.  We have used the corrected version of the Indian pines and Salinas dataset having the number of bands 200 and 204 respectively. 
	The Pavia University dataset uses 103 bands.  The pre-processing of the datasets is the same as done in \cite{santara2017bass}, following the code available in\cite{hyperspectral2016KGPML}. For any dataset, its pixel values are scaled to $[0,1]$ using the expression $(x-\min(x))/(\max(x)-\min(x))$, where, $x$ is a pixel value. The $\max$ and $\min$ are computed over the entire HSI. The data are then mean normalized across each channel by subtracting channel-wise means. The datasets are partitioned into training and test datasets. For band selection, only the training datasets are fed to the model. For measuring performances both training and test datasets are used. For splitting the datasets into training and test subsets, we drop the pixels of the unknown land-cover type. Let $\mathbf{X}$ be the set of pixels with known land-cover type. To obtain the training and test sets, let us divide $\mathbf{X}$ into two subsets $\mathbf{A}$ and $\mathbf{B}$ such that $\mathbf{A}\bigcup \mathbf{B}=\mathbf{X}$,  $\mathbf{A}\bigcap \mathbf{B}=\phi$, and $\mathbf{A}$ and $\mathbf{B}$ contain, respectively,  $25\%$ and $75\%$  pixels of $\mathbf{X}$. We use $\mathbf{A}$ as the test set. Note that, both the datasets suffer from the class imbalance problem. To avoid the learning difficulty raised by class imbalance, in the training set, we consider the same number of instances from each class. For this, from the subset $\mathbf{B}$, we randomly select (without replacement) $200$ pixels per class. If a class has less than $200$ instances in $\mathbf{B}$,  we oversample the class by synthetic minority oversampling  technique (SMOTE) \cite{chawla2002smote} to gather $200$ points. For band selection also, we use the same neural network (Fig. \ref{fig:network1}) with the number of hidden layers, $n_H=3$.  The numbers of hidden nodes in the three hidden layers are $500, 350,$ and  $150$ respectively. Here the number of input nodes of the MLP is equal to the number of bands ($P$). The network weights and the gate parameters $\lambda_{j}$s are initialized in the same way as done. For all experiments of the current sub-section, $\alpha_{1}$ and $\alpha_{2}$ of the error functions in Equations (\ref{eq:error_cl_band}) and (\ref{eq:overall_error}) are set as $5$ and $1$ respectively. The total number of iterations for training the network is set to $50000$. The rest of experimental settings are kept same as the previously mentioned experiment with the two data sets. The number of training instances of Indian pines and Salinas data set 
 is $3200$ and that of Pavia university is $1800$. Both of the  number of training instances, $n$ are high. Computation of the $E_{sammons}$ in (\ref{eq:sammons}) would involve computing $(3200)^{2}$ or $(1800)^{2}$ distances. Adding $E_{sammons}$ to the overall loss function would cause very intensive computation at each iteration. So, instead of  $E_{sammons}$, its proposed approximation $E_{struct}$ defined in (\ref{eq:struct}) is used.  In (\ref{eq:struct}), $|S_{t}|$ is taken as $100$. Varying the value of $\beta$ in (\ref{eq:overall_error}), we analyse its effect on the OCA, SS, NMI, ARI, and JI. We compute SS, NMI, ARI, and JI as described in Subsec. \ref{subsec:chap4_expt_general}.  We also use the same clustering algorithm with the same settings as used in Subsec. \ref{subsec:chap4_expt_general}.
	
	Tables \ref{tab:perf_indian_pines_train} and \ref{tab:perf_indian_pines_test} summarize the comparative results of  FSMLP\textsubscript{struct} with FSMLP and other band selection methods, ICA, F score, and mutual information based filter methods on the training and test datasets of Indian pines respectively.
	\begin{table}[!tb]
		\caption{Performance comparison for Indian Pines training set, with the number of bands= $70$.}\label{tab:perf_indian_pines_train}
		\centering
		\begin{tabular}{|c|c c c c c|}
			\hline
			Method & SS & NMI & ARI & JI & OCA \\
			\hline
			ICA & 0.6982 & 0.5437 & 0.3071 & 0.3335 & 0.7497 \\
			F Score & 0.5325 & 0.5567 & 0.3015 & 0.0704 & 0.7978 \\
			Mutual Info & 0.1966 & 0.5633 & 0.3232 & 0.2493 & 0.8600 \\
			FSMLP & 0.0162 & 0.7994 & 0.6530 & 0.6827 & 0.9157 \\
			FSMLP\textsubscript{struct}, $\beta$=2 & 0.0153 & 0.8027 & 0.6588 & 0.6894 & 0.9165 \\
			FSMLP\textsubscript{struct}, $\beta$=5 & 0.0140 & 0.8051 & 0.6627 & 0.6945 & 0.9155 \\
			FSMLP\textsubscript{struct}, $\beta$=20 & 0.0107 & 0.8188 & 0.6953 & 0.7216 & 0.9153 \\
			FSMLP\textsubscript{struct}, $\beta$=50 & \textbf{0.0071} & \textbf{0.8400} & \textbf{0.7404} & \textbf{0.7549} & \textbf{0.9179} \\
			\hline
		\end{tabular}
	\end{table}
	\begin{table}[!tb]
		\caption{Performance comparison for Indian Pines test set, with the number of bands= $70$.}\label{tab:perf_indian_pines_test}
		\centering
		\begin{tabular}{|c|c c c c c|}
			\hline
			Method & SS & NMI & ARI & JI & OCA \\
			\hline
			ICA & 0.6748 & 0.5360 & 0.3052 & 0.2983 & 0.6175 \\
			F Score & 0.5457 & 0.5518 & 0.2951 & 0.1643 & 0.6738 \\
			Mutual Info & 0.2041 & 0.5787 & 0.3283 & 0.2395& 0.7298 \\
			FSMLP & 0.0158 & 0.8729 & 0.7949 & 0.8109 & 0.7871 \\
			FSMLP\textsubscript{struct}, $\beta$=2 & 0.0148 & 0.8831 & 0.8210 & 0.8384 & 0.7871 \\
			FSMLP\textsubscript{struct}, $\beta$=5 & 0.0137 & 0.8717 & 0.7881 & 0.8017 & 0.7881 \\
			FSMLP\textsubscript{struct}, $\beta$=20 & 0.0102 & 0.8896 & 0.8346 & \textbf{0.8507} & \textbf{0.7882} \\
			FSMLP\textsubscript{struct}, $\beta$=50 & \textbf{0.0066} & \textbf{0.8926} & \textbf{0.8360} & 0.8505 & 0.7850 \\
			\hline
		\end{tabular}
	\end{table}
	Similarly, Table \ref{tab:perf_pavia_train} and \ref{tab:perf_pavia_test} summarize the comparative results on the training and test datasets of Pavia university. 
	\begin{table}[!tb]
		\caption{Performance comparison for Pavia University training set, with the number of bands= $35$.}\label{tab:perf_pavia_train}
		\centering
		\begin{tabular}{|c|c c c c c|}
			\hline
			Method & SS & NMI & ARI & JI & OCA \\
			\hline
			ICA & 0.3231 & 0.6941 & 0.5858 & 0.6340 & 0.8433 \\
			F Score & 0.2386 & 0.6451 & 0.4654 & 0.4748 & 0.8233 \\
			Mutual Info & \textbf{0.1387} & 0.8905 & 0.8769 & 0.8871 & 0.8717 \\
			FSMLP & 0.1686 & 0.9072 & 0.8925 & 0.8863 & 0.9213 \\
			FSMLP\textsubscript{struct}, $\beta$=1 & 0.1625 & 0.9169 & 0.9014 & 0.8952 & 0.9213 \\
			FSMLP\textsubscript{struct}, $\beta$=1.5 & 0.1616 & 0.9233 & 0.9112 & 0.9033 & 0.9212 \\
			FSMLP\textsubscript{struct}, $\beta$=2 & 0.1592 & \textbf{0.9243} & \textbf{0.9121} & \textbf{0.9044} & 0.9213 \\
			FSMLP\textsubscript{struct}, $\beta$=2.5 & 0.1590 & 0.9198 & 0.9056 & 0.8993 & \textbf{0.9218} \\
			\hline
		\end{tabular}
	\end{table}
	\begin{table}[!tb]
		\caption{Performance comparison for Pavia University test set, with the number of bands= $35$.}\label{tab:perf_pavia_test}
		\centering
		\begin{tabular}{|c|c c c c c|}
			\hline
			Method & SS & NMI & ARI & JI & OCA \\
			\hline
			ICA & 0.3215 & 0.5533 & 0.3777 & 0.2082 & 0.7573 \\
			F Score & 0.3350 & 0.5091 & 0.3126 & 0.0886 & 0.7222 \\
			Mutual Info & \textbf{0.1433} & 0.8911 & 0.8814 & 0.8990 & 0.7785 \\
			FSMLP & 0.1584 & 0.9001 & 0.8930 & 0.9084 & 0.8628 \\
			FSMLP\textsubscript{struct}, $\beta$=1 & 0.1568 & 0.9156 & 0.9131 & 0.9254 & 0.8650 \\
			FSMLP\textsubscript{struct}, $\beta$=1.5 & 0.1573 & 0.9175 & 0.9149 & 0.9270 & \textbf{0.8672} \\
			FSMLP\textsubscript{struct}, $\beta$=2 & 0.1556 &  \textbf{0.9189} & \textbf{0.9166} & \textbf{0.9279}  & 0.8666 \\
			FSMLP\textsubscript{struct}, $\beta$=2.5 & 0.1562 & 0.9168 & 0.9143 & 0.9261 & 0.8648 \\
			\hline
		\end{tabular}
	\end{table}

	In this experiment, we have fixed the number of selected bands $Q$, approximately to $35\%$ of the original number of bands $P$. So, The number of selected bands is $70$ for Indian pines and 
	it is $35$ for Pavia University. Tables \ref{tab:perf_indian_pines_train} and \ref{tab:perf_indian_pines_test} record the values of the structure-preserving indices and classification scores on Indian pines  for different $\beta$ values in FSMLP\textsubscript{struct} ($\beta$ values in Equation (\ref{eq:overall_error})). The considered $\beta$s for Indian pines, 
	are $2, 5, 20,$ and $50$. Note here that, FSMLP is basically FSMLP\textsubscript{struct} with $\beta=0$. We observe in Tables \ref{tab:perf_indian_pines_train} and \ref{tab:perf_indian_pines_test} that both for training and test datasets as the value of $\beta$ increases for FSMLP\textsubscript{struct} (in the last five rows of the corresponding Tables) the value of SS becomes smaller. A similar trend  is also observed for the Pavia University data set (here, $\beta$ varies as $1,1.5,2,$ and $2.5$) for training (Table \ref{tab:perf_pavia_train}) and test (Table \ref{tab:perf_pavia_test}) sets.  
	
	For the Pavia university dataset, we have set the values of $\beta$ in FSMLP\textsubscript{struct} as $1,1.5,2,$ and $2.5$. Unlike Indian pines  
	for Pavia university, we restrict the $\beta$s to lower values. This is due to the fact that the number of selected bands for Pavia university is $35$ and that for Indian pines 
	is $70$. Lesser the number of bands, the lesser the importance ($\beta$) to be given to our structure preserving regularizer in Equation (\ref{eq:struct}) to obtain a desired balance between classification and clustering performance. Table \ref{tab:perf_indian_pines_train} which contains the results for the Indian pines training data, clearly shows that both FSMLP and FSMLP\textsubscript{struct} are better than ICA, F-score based, mutual information based methods in all four structure preserving metrics as well as in terms of the OCA. In Table \ref{tab:perf_indian_pines_train} we observe that with increasing values of $\beta$ there is a consistent improvement in the values of the four structure preserving metrics while the values of OCAs retain approximately at $91\%$. The results shown in Table \ref{tab:perf_indian_pines_test} for the Indian pines test set also show that FSMLP and FSMLP\textsubscript{struct} perform better in terms of all the five metrics than the other three methods. Also with an increase in $\beta$ all the structure-preserving metrics improve for FSMLP\textsubscript{struct}, except the value of JI slightly decreases when $\beta$ goes to $50$ from $20$. The classification metric OCA is around $78\%$ with bands selected by FSMLP\textsubscript{struct} for different choices of $\beta$s. 
	
	It is notable here that the test set is completely unseen in the process of band selection, yet the selected bands for the proposed method is providing fairly good results for structure preservation as well as for classification. As observed from Table \ref{tab:perf_pavia_train} and Table \ref{tab:perf_pavia_test} for Pavia university training and test sets respectively, the lowest (best) SS value among all the comparing methods is achieved by mutual information based filter method.  However, for the other four metrics i.e. NMI, ARI, JI and OCA; FSMLP and FSMLP\textsubscript{struct} show better values. In the case of the Pavia university dataset with increasing $\beta$; NMI, ARI, and JI are not consistently increasing but the results indicate that, it is possible to find a $\beta$, (here $\beta=2$) where the structures are preserved better maintaining a good classification score.  
	Table \ref{tab:perf_salinas_train} and \ref{tab:perf_salinas_test} summarize the comparative results of training and test datasets of Salinas. 
	
	\begin{table}[!h]
		\caption{Performance comparison for Salinas training set, with the number of bands= $70$.}\label{tab:perf_salinas_train}
		\centering
		\begin{tabular}{|c|c c c c c|}
			\hline
			Method & SS & NMI & ARI & JI & OCA \\
			\hline
			ICA	&	0.7159	&	0.7066	&	0.5385	&	0.5111	&	0.9325	\\
			F Score	&	0.0167	&	0.9420	&	0.9307	&	0.9291	&	0.9650	\\
			Mutual Info	&	0.0902	&	0.9207	&	0.8775	&	0.8860	&	0.9628	\\
			FSMLP	&	0.0072	&	0.9672	&	0.9589	&	0.9628	&	0.9654	\\
			FSMLP\textsubscript{struct}, $\beta=2$	&	0.0078	&	0.9678	&	0.9603	&	0.9640	&	\textbf{0.9658}	\\
			FSMLP\textsubscript{struct}, $\beta=5$	&	0.0072	&	0.9691	&	0.9626	&	0.9659	&	0.9648	\\
			FSMLP\textsubscript{struct}, $\beta=20$	&	0.0055	&	\textbf{0.9710}	&	\textbf{0.9657}	&	\textbf{0.9683}	&	0.9646	\\
			FSMLP\textsubscript{struct}, $\beta=50$	&	\textbf{0.0039}	&	0.9688	&	0.9630	&	0.9659	&	0.9649	\\
			\hline
		\end{tabular}
	\end{table}
	\begin{table}[!h]
		\caption{Performance comparison for Salinas test set, with the number of bands= $70$.}\label{tab:perf_salinas_test}
		\centering
		\begin{tabular}{|c|c c c c c|}
			\hline
			Method & SS & NMI & ARI & JI & OCA \\
			\hline
			ICA	&	0.7027	&	0.6297	&	0.4144	&	0.3490	&	0.8670	\\
			F Score	&	0.0230	&	0.8584	&	0.7483	&	0.7615	&	0.9081	\\
			Mutual Info	&	0.0820	&	0.8791	&	0.7950	&	0.8230	&	0.8950	\\
			FSMLP	&	0.0078	&	0.9368	&	0.9183	&	0.9265	&	\textbf{0.9077}	\\
			FSMLP\textsubscript{struct}, $\beta=2$	&	0.0084	&	0.9392	&	0.9231	&	0.9305	&	0.9076	\\
			FSMLP\textsubscript{struct}, $\beta=5$	&	0.0079	&	\textbf{0.9413}	&	\textbf{0.9275}	&	\textbf{0.9340}	&	0.9069	\\
			FSMLP\textsubscript{struct}, $\beta=20$	&	0.0061	&	0.9366	&	0.9175	&	0.9258	&	0.9065	\\
			FSMLP\textsubscript{struct}, $\beta=50$	&	\textbf{0.0044}	&	0.9342	&	0.9136	&	0.9229	&	0.9073	\\
			\hline
		\end{tabular}
	\end{table}
	We note from Tables \ref{tab:perf_salinas_train} and \ref{tab:perf_salinas_test} that, for the Salinas dataset, FSMLP and FSMLP\textsubscript{struct} are better than the other three methods in all the five metrics used. All four structure preserving metrics scores of FSMLP\textsubscript{struct} are better than or comparable to FSMLP keeping the classification score OCA at approximately $96\%$ for the training dataset and $90\%$ for the test dataset. Tables \ref{tab:perf_salinas_train} and \ref{tab:perf_salinas_test} reveal that when $\beta$ is increased from $0$ to $2$, the value of SS is increased however, from $\beta=2$ to $\beta=50$ onward, the values of SS are decreased. The exceptions for the Salinas dataset, while increasing $\beta$ from $0$ to $2$ is possibly due to the fact that we do not use the entire training data in Equation (\ref{eq:struct}) and  use of $|S_{t}|=100$ in Equation (\ref{eq:struct}) is not adequate to capture the structure of the data faithfully for  the Salinas dataset. As discussed earlier, setting the value of $|S_{t}|$ is crucial for approximating Equation (\ref{eq:sammons}) with Equation (\ref{eq:struct}). We have set $|S_{t}|=100$ for all three datasets empirically. However, choosing an optimum value of $|S_{t}|$ for each dataset is expected to avoid the occurred exceptions.
	
	 As we increase the value of $\beta$, there is more stress to reduce the loss function Equation (\ref{eq:struct}). In most cases, increasing $\beta$ results in a drop in SS. This  clearly suggests that the loss function Equation (\ref{eq:struct}) that we use, is a computationally efficient substitute for the original SS defined in Equation (\ref{eq:sammons}). 
	 
\begin{figure}
	 	\centering
	 	\begin{subfigure}{.48\textwidth}
	 		\centering
	 		\includegraphics[width=.98\linewidth]{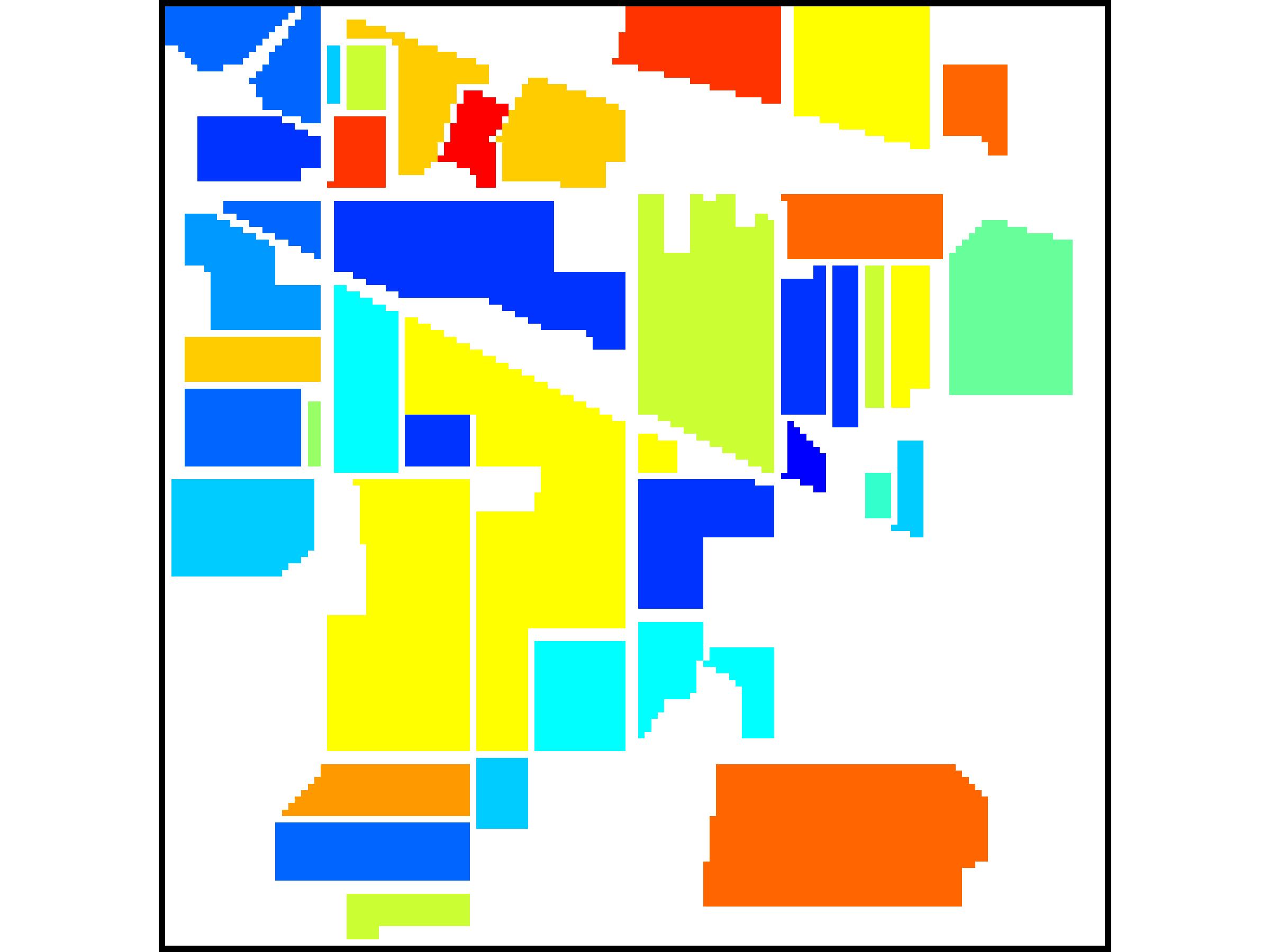}
	 		\caption{ }
	 		\label{fig:ip_gt}
	 	\end{subfigure}
	 	\begin{subfigure}{.48\textwidth}
	 		\centering
	 		\includegraphics[width=.98\linewidth]{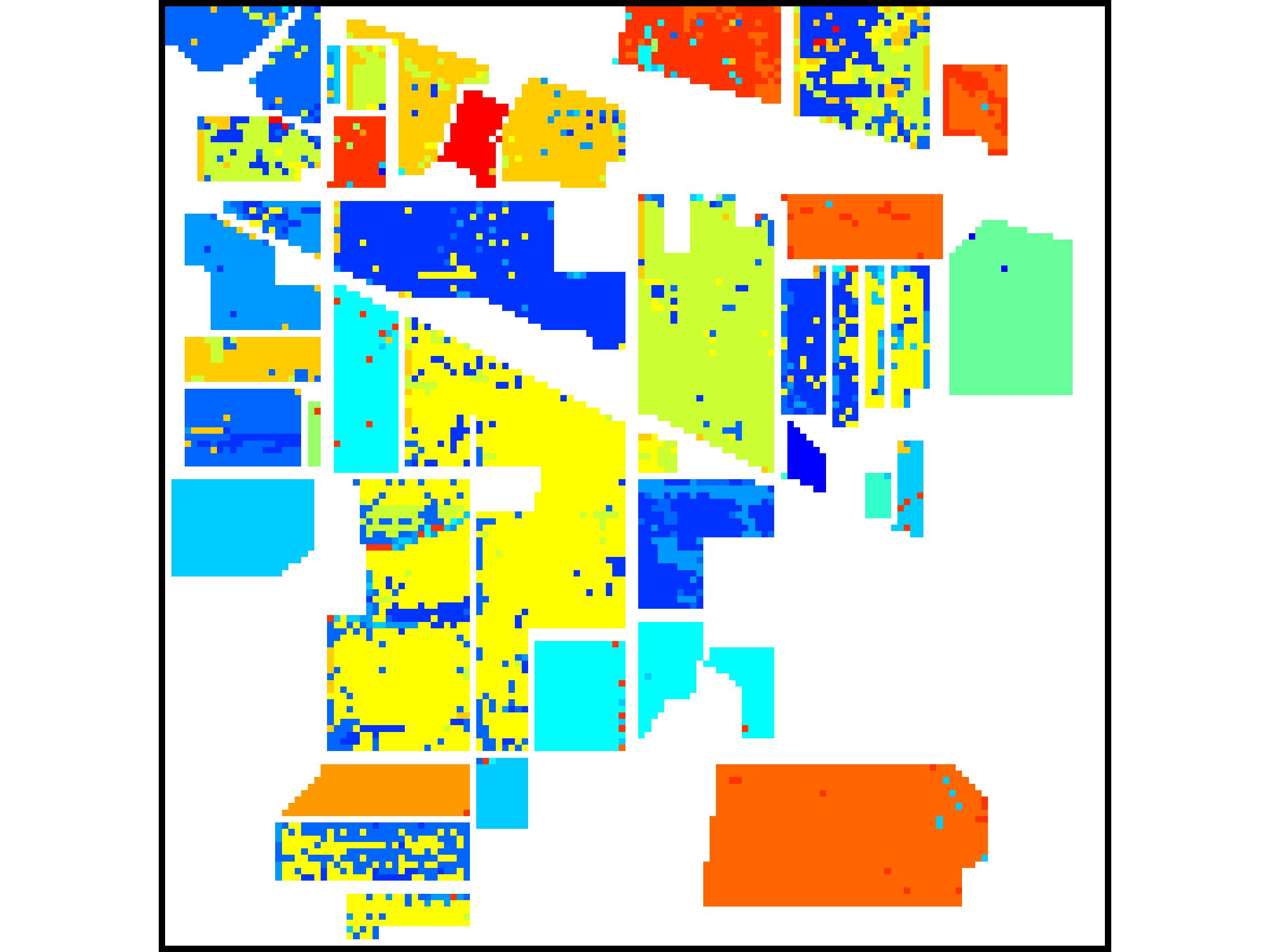}
	 		\caption{ }
	 		\label{fig:ip_FSMLP}
	 	\end{subfigure}\\
	 	\begin{subfigure}{.48\textwidth}
	 		\centering
	 		\includegraphics[width=.98\linewidth]{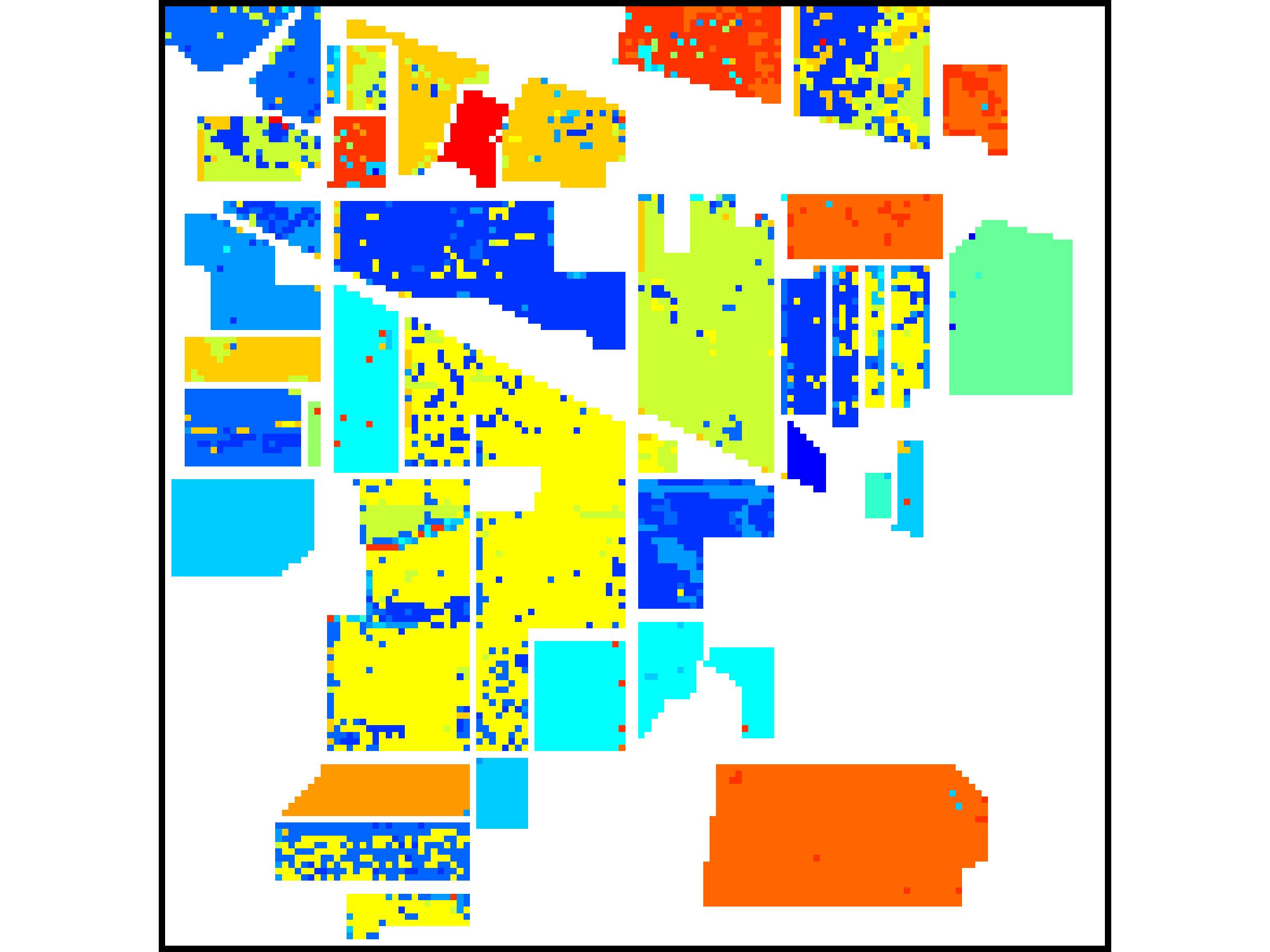}
	 		\caption{ }
	 		\label{fig:ip_FSMLP_struct_20}
	 	\end{subfigure}
	 	\begin{subfigure}{.48\textwidth}
	 		\centering
	 		\includegraphics[width=.98\linewidth]{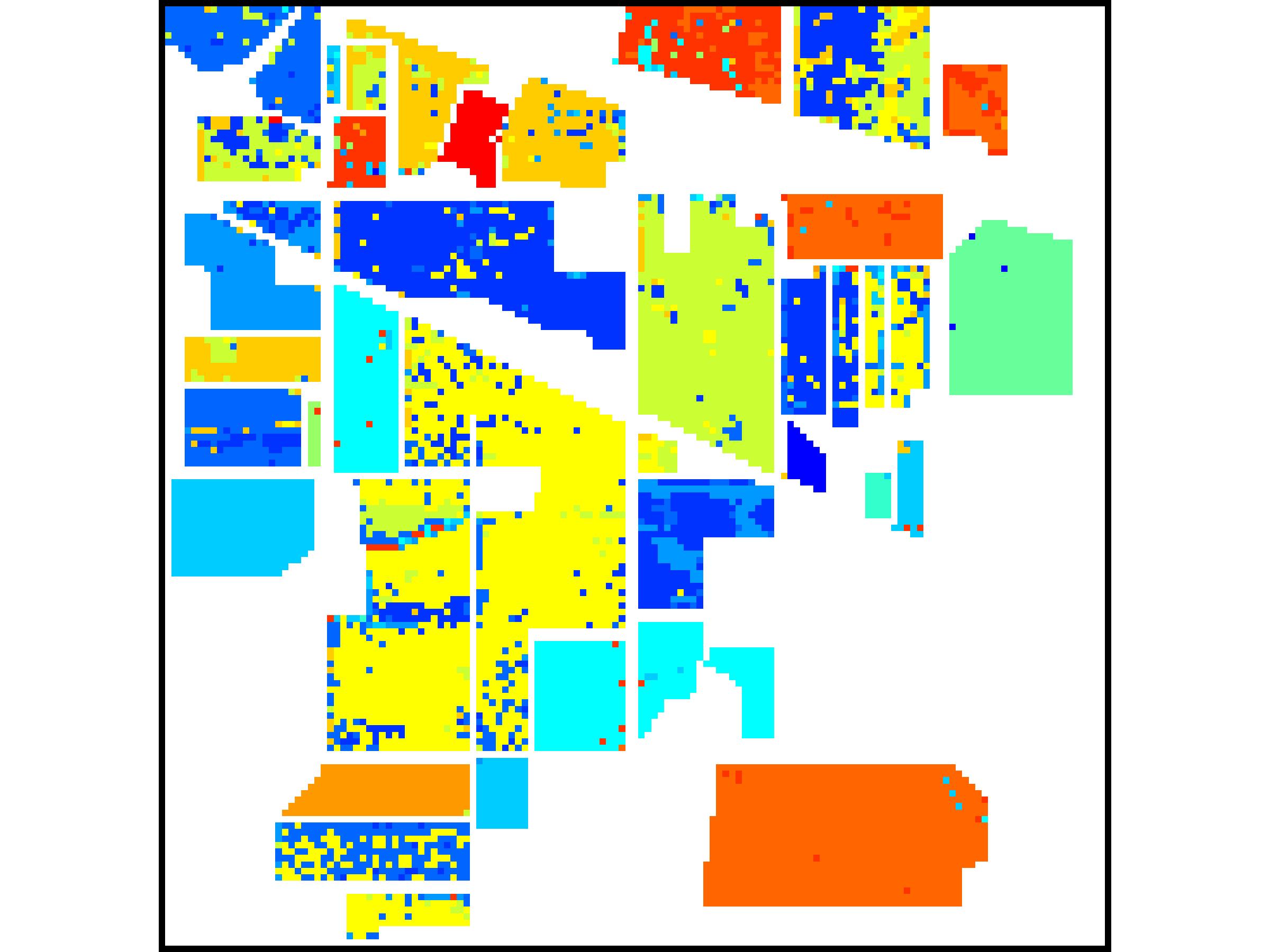}
	 		\caption{ }
	 		\label{fig:ip_FSMLP_struct_50}
	 	\end{subfigure}	
	 	\caption{Thematic maps resulting from classifications (SVM) of the Indian pines dataset with $70$ bands selected by FSMLP\textsubscript{struct} considering (a) $\beta=0$, (b) $\beta=20$, (c) $\beta=50$.}
	 	\label{fig:thematic_map_ip_betas}
	 \end{figure}

	 We have included results of the thematic maps (Fig. \ref{fig:thematic_map_ip_betas}) and it reveals that our proposed method is capable of selecting useful bands that can broadly capture the land cover types.
	 Figure \ref{fig:thematic_map_ip_betas} illustrates thematic maps of the entire region captured in the Indian pines dataset. Figure \ref{fig:ip_gt} shows ground truth labels. Figures \ref{fig:ip_FSMLP}, \ref{fig:ip_FSMLP_struct_20}, \ref{fig:ip_FSMLP_struct_50} are thematic maps of the Indian pines data set using the class labels obtained from the SVM classifier trained on the considered training set  represented with $70$ bands selected by FSMLP\textsubscript{struct} with $\beta=0$, i.e., by the method FSMLP, and FSMLP\textsubscript{struct} considering $\beta=20$, and $\beta=50$, respectively. 
	 Figure \ref{fig:thematic_map_ip_betas} ensures that even with the increasing stress on the structure-preserving regularizer $E_{struct}$, our proposed band selection method FSMLP\textsubscript{struct} is able to select bands that maintain a  good  land cover classification performance. 
	 
	\section{Conclusion and Discussions}\label{sec:chap4_conclusion}
	To the best of our knowledge, a feature selection method that simultaneously cares about class discrimination and structure preservation is not available in the literature. In this study, we have tried to bridge this gap by proposing a neural network-based feature selection method that focuses both on class discrimination and structure preservation. To learn the proposed system, we use Sammon's stress as a regularizer to the classification loss. For datasets having a large number of instances, the computational overhead associated with Sammon's stress is very high. Consequently, as the structure-preserving regularizer, we use Sammon's stress computed based on a sample of the original data (using dynamic sampling on each iteration during the adaptive gradient descent based learning). Using this regularizer in the experiments with datasets having a large number of instances, we have demonstrated that this regularizer is an effective and computationally efficient implementation of Sammon's stress based structure-preserving regularizer. Our proposed feature selection scheme is generic. So we have investigated its effectiveness on datasets commonly used for assessing classifiers as well as for a specialized case: band selection in hyperspectral images (HSI). We have applied the feature selection scheme to five real-world datasets which are commonly used typically for assessing classification. In the context of band selection, we have applied our method to three well-known HSI datasets and compared performances with three other band selection methods. Based on our experiments, we conclude that the proposed feature selection method is able to produce reasonably good  classification and clustering scores in the majority of the data sets, proving that the proposed method is capable of selecting a subset of features that is good both for classification and clustering. Our scheme provides a mechanism to control the number of selected features. The proposed method is easily extendable to other networks like Radial Basis Function (RBF) network.
	
	\bibliography{references_FS}      
\end{document}